\documentclass[12pt]{article}

\usepackage{amssymb,amsmath}
\usepackage{graphicx}
\usepackage{float}
\usepackage{caption}
\usepackage{subcaption}
\usepackage[export]{adjustbox}
\usepackage{supertabular}
\usepackage{hyperref}
\usepackage{xcolor}
\usepackage{geometry}
\usepackage{lineno}

\usepackage{cite}

\title{Transport-Related Surface Detection with Machine Learning:\\Analyzing Temporal Trends in Madrid and Vienna}

\author{
Miguel Ureña Pliego$^{1}$,
Rubén Martínez Marín$^{1}$,
Nianfang Shi$^{2}$,\\
Takeru Shibayama$^{2}$,
Ulrich Leth$^{2}$,
Miguel Marchamalo Sacristán$^{1}$\\[10pt]
\small $^{1}$Department of Land Morphology and Engineering, Civil Engineering School, \\ 
\small Universidad Politécnica de Madrid, C. del Prof. Aranguren 3, 28040 Madrid, Spain\\
\small $^{2}$Institut für Verkehrswissenschaften, \\
\small TU Wien, Karlsplatz 13/230, A-1040 Wien, Austria \\
\small Contact: miguel.urena@upm.es
}

\date{08/2024}

\begin{document}

\maketitle

\begin{abstract}
This study explores the integration of machine learning into urban aerial image analysis, with a focus on identifying infrastructure surfaces for cars and pedestrians and analyzing historical trends. It emphasizes the transition from convolutional architectures to transformer-based pre-trained models, underscoring their potential in global geospatial analysis. A workflow is presented for automatically generating geospatial datasets, enabling the creation of semantic segmentation datasets from various sources, including WMS/WMTS links, vectorial cartography, and OpenStreetMap (OSM) overpass-turbo requests. The developed code allows a fast dataset generation process for training machine learning models using openly available data without manual labelling. Using aerial imagery and vectorial data from the respective geographical offices of Madrid and Vienna, two datasets were generated for car and pedestrian surface detection. A transformer-based model was trained and evaluated for each city, demonstrating good accuracy values. The historical trend analysis involved applying the trained model to earlier images predating the availability of vectorial data 10 to 20 years, successfully identifying temporal trends in infrastructure for pedestrians and cars across different city areas. This technique is applicable for municipal governments to gather valuable data at a minimal cost.
\end{abstract}


\vspace{0.5em}

This document is a preprint. Peer reviewed version available under \url{https://doi.org/10.1016/j.rsase.2025.101503} \\

\newpage 

\noindent\textbf{Highlights:}
\begin{itemize}
  \item General purpose semantic segmentation models do not provide accurate results when working with aerial imagery.
  \item Large training datasets for machine learning vision models were created using openly available aerial imagery and vectorial data from OpenStreetMap and official sources.
  \item The dataset creation did not require manual human labelling and the code created for this study is easy to use and fast.
  \item The vision models detected street, sidewalk, parking, and other classes with medium to high accuracy.
  \item The accuracy and usability of the vision models were tested in the cities of Madrid and Vienna, detecting temporal tendencies in the evolution of transportation-related surfaces.
  \item It was possible to detect a different tendency in the evolution of the amount of parking and road surface, with a decrease in the centre and an increase in the suburbs of the city of Madrid.
\end{itemize}

\newpage 

\noindent\textbf{Graphical Abstract:}
\begin{figure}[H]
\centering
\includegraphics[width=\linewidth]{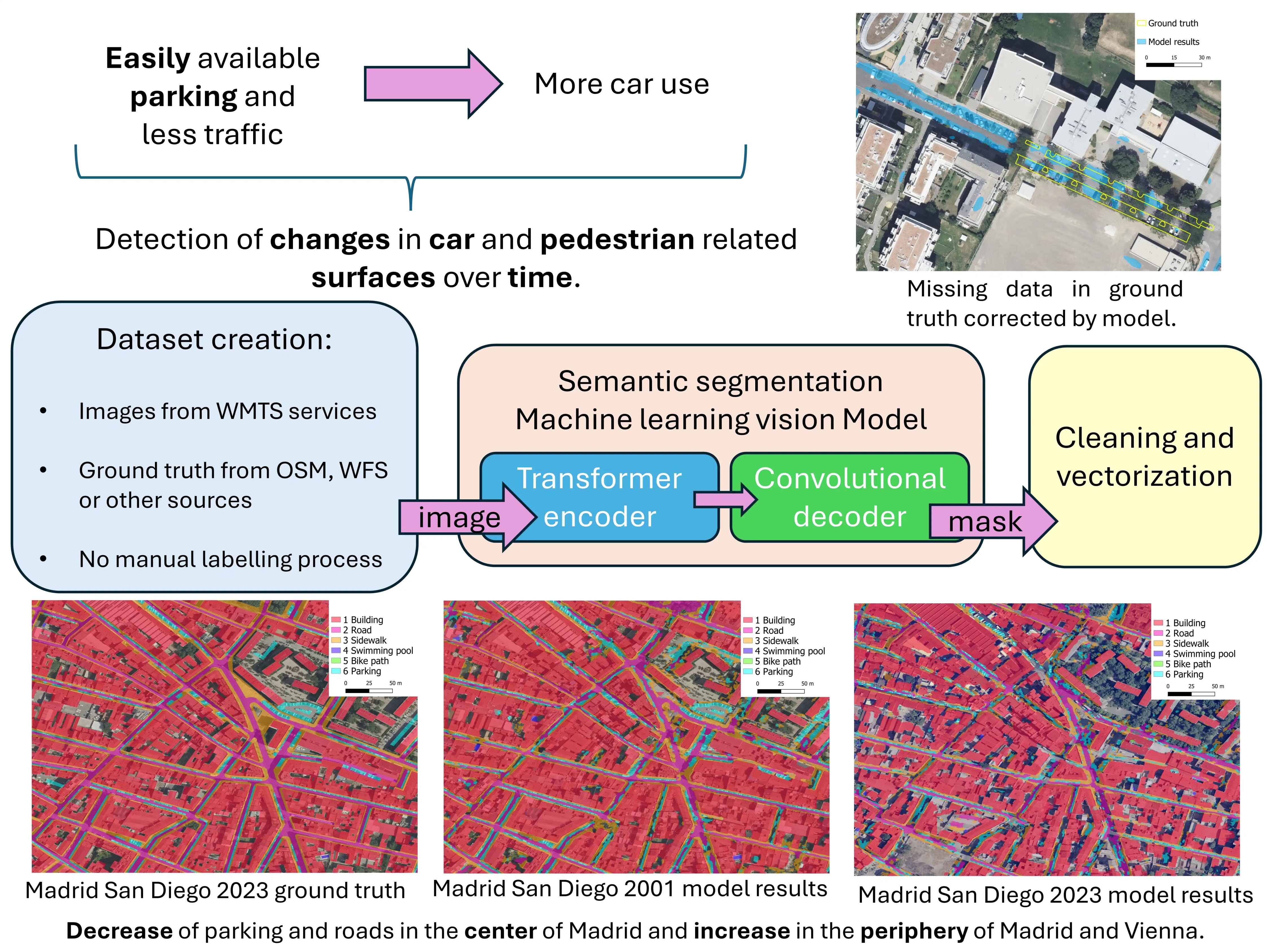}
\caption*{Graphical abstract illustrating the workflow and results.}
\end{figure}

\newpage 

\section{Introduction and Objectives}

Satellite imagery has been globally accessible for the past two decades, while high-resolution aerial imagery from airplanes has been available since the mid-20th century. Both current and historical images are freely accessible for research via services like Google Earth Engine \cite{gorelick_google_2017} and national or local cartography providers. 

Vectorial cartography is created by tracing aerial imagery. OpenStreetMap (OSM), the largest crowd-sourced project, relies on this method for global vectorial cartography \cite{haklay_openstreetmap_2008}. However, this manual process is labor-intensive, resulting in incomplete data, especially in non-common categories like on-street parking spaces. Historical urban analysis is challenging due to scarce, less detailed historical cartography. \\

Aerial imagery is widely available, but vectorial cartography data is more accessible in developed regions with Spatial Data Infrastructures (SDIs) like the European Union’s INSPIRE scheme, which standardizes geospatial datasets \cite{noauthor_directive_2007, minghini_comparing_2019}. However, categories may lack clear definitions. For instance, parking falls under RoadService without further subdivisions. 
In major European cities, openly accessible vector cartography following the INSPIRE scheme usually undergo  
annual updates \cite{noauthor_city_nodate, noauthor_stadtvermessung_nodate,
paris_city_council_paris_nodate, 
noauthor_luftbilder_nodate, 
ayuntamiento_de_madrid_geoportal_nodate}. 

OSM, as it is maintained by volunteers, does not follow the INSPIRE scheme but offers diverse geospatial data based on local needs \cite{minghini_comparing_2019}. OSM enables collaboration for rapid cartography, especially useful for humanitarian mapping post-disasters \cite{herfort_evolution_2021}. OSM has 88931 
different keys, but the dataset is not comprehensive, and the absence 
of numerous objects can be anticipated. The data's quality varies, posing challenges for scientific studies \cite{minghini_openstreetmap_2019, haklay_how_2010}. \\

Advancements in computer vision and object detection \cite{zou_object_2023} have enabled semantic segmentation and object detection in aerial imagery, useful across many fields \cite{yuan_review_2021}. Given the challenges of training general models and the availability of high-resolution imagery and cartography, it is important to establish a workflow for creating specific training datasets and training vision models for object detection and semantic segmentation in aerial imagery. \\

Analyzing surfaces dedicated to cars and pedestrians is crucial for evaluating car use in neighborhoods \cite{nurul_habib_integrating_2012}. Although government and OSM data exist, they often lack quality and historical data are often not available. Machine learning can address these gaps. \\

Our research aims to create a workflow for developing datasets from global aerial imagery and vectorial cartography, applied to Madrid and Vienna for car and pedestrian surfaces. This workflow accelerates dataset creation, the most time-consuming part of a machine learning project. We want to demostrate the possibility of adapting and fine tuning existing semantic segmentation models for gathering data on urban infrastructure, especially for classes like parking spaces and demonstrate the viability of extracting historical trends using this method. 

\section{Background}

Machine learning has seen extensive application in remote sensing 
and object detection over the last decade 
\cite{zou_object_2023}. 
In the earlier stages, 
traditional statistical methods were prevalent for tasks 
like automatic road extraction \cite{mena_state_2003}. 
However, machine learning methods have surpassed traditional approaches 
in recent years \cite{zou_object_2023}. Various datasets and benchmarks for semantic segmentation 
tasks related to building detection, land use and cover extraction, 
and road geometry are available \cite{zamir_isaid_2019,
helber_introducing_2018, 
mnih_machine_2013, 
marmanis_semantic_2016}.

Machine learning models working with aerial imagery in the optical 
spectrum have been developed for diverse applications, including 
land use detection \cite{talukdar_land-use_2020, 
nasiri_land_2022}, building detection 
\cite{zhu_map-net_2021, 
dong_review_2024, 
shao_brrnet_2020, 
liu_building_2019}, 
and road geometry extraction 
\cite{mattyus_deeproadmapper_2017, 
zhang_road_2018}.

In most cases, these machine learning models are based on image 
convolutions. They play a vital role in annotating maps 
\cite{jiao_machine_2018, 
abdurakhmonov_advances_2023} and exhibit high 
accuracy when tested in environments similar to the training data, 
but a low generability to unknown environments. 
Additionally, machine learning models can be utilized to correct 
OpenStreetMap (OSM) data 
\cite{vargas-munoz_openstreetmap_2021}.

Researchers have explored the labeling and creation of more diverse  
datasets to enhance the generability of aerial imagery 
segmentation models across diverse global environments 
\cite{maggiori_can_2017}. However, addressing 
this challenge remains an ongoing effort. Recent advancements include 
the development of general purpose models that leverage 
transformer-based machine learning architectures for geospatial 
analysis \cite{wu_samgeo_2023}. 
A commercially available 
plugin for QGIS is now accessible 
\cite{aszkowski_deepness_2023} to run such models 
directly.

Datasets like GBSS 
\cite{hu_gbssglobal_2024} by Google and 
ML Building Footprints by Microsoft 
\cite{microsoft_globalmlbuildingfootprints_2024} provide 
building footprints worldwide. Both datasets where 
created with machine learning models trained with OSM data. 
The datasets may not have a very high quality  
\cite{gonzales_building-level_2023}, especially 
when precise annotations are required or when models are tested in 
regions outside the Western world. \\

While there are methods for detecting vehicles and monitoring parking space occupancy, they do not specifically extract the parking and road surface from individual images as it is intended in this study. These methods focus on detecting the cars rather than the parking surface itself. There are techniques involving surveillance cameras \cite{amato_deep_2016} and UAV high-resolution images, with resolutions ranging from 5 to 1.5 cm per pixel \cite{audebert_segment-before-detect_2017, kumar_deep_2022} and 
multispectral high resolution aerial imagery \cite{merkle_semantic_2019}. Our objective is to create an inventory using openly available data with much lower resolution and to segment surfaces independently of the presence of cars.

Additionally, there is a method using satellite imagery that statistically evaluates image histograms to extract parking space occupancy without detecting individual cars \cite{drouyer_parking_2020}. However, this method is not suitable for our purpose.

Another intriguing area of research involves utilizing OpenStreetMap (OSM) data to generate datasets for tasks similar to ours. Some studies have created datasets to analyze transportation infrastructure, particularly parking and road surfaces \cite{henry_citywide_2021, hellekes_parking_2023}, aiming to develop a comprehensive parking space inventory. These studies achieved good results with Intersection over Union (IoU) values exceeding 60. In these cases, OSM data were refined and manually annotated, which is both time-consuming and expensive and the generability of the model has not been 
evaluated, so applications in areas different from the training set are not suitable.

\subsection{The relevance of road and parking surface in encouraging car usage}

When studying the reasons people commute by car, research has long shown \cite{abrahamse_factors_2009, carse_factors_2013} that the main factors are related to self-interest. Specifically, people choose to drive because it is the most convenient way for them to commute. Additionally, individuals who are more morally aware tend to avoid using cars. Those who do choose to drive often feel more guilt or responsibility regarding the environmental and social problems associated with car-centric behavior \cite{abrahamse_factors_2009}.

The amount of car-related infrastructure is therefore one of the main factors encouraging car use and diverting financing away from more sustainable alternatives \cite{mattioli_political_2020}. Parking space availability, in particular, has long been recognized as one of the most critical factors influencing the choice of travel by car \cite{feeney_review_1989, christiansen_parking_2017, mccahill_effects_2016, pandhe_parking_2012}. This is acknowledged by the European Commission in its technical guide for parking policy \cite{tom_rye_parking_2022}. Parking space availability in urban environments can be indirectly inferred from the amount of paved surface \cite{litman_why_2011} or directly evaluated by segmenting the area dedicated to that purpose from aerial images.

For another important factor in the modal split, the surface dedicated to active modes in already developed urban areas competes directly with surface dedicated to cars \cite{gerike_built_2021}. The relationship between surfaces dedicated to cars and pedestrians has a major and direct influence on walkability and the choice of active modes instead of driving \cite{gerike_built_2021}. In new urban developments, dedicating too much surface to cars makes distances and destinations farther apart, encouraging urban sprawl and car use \cite{alshammari_compactness_2022, glaeser_chapter_2004}. New approaches to street design consider reallocating urban space from cars to pedestrians and cyclists, redesigning streets to consider the needs of all users and encourage sustainable behavior \cite{halpern_road_2022, rodriguez-valencia_understanding_2021, schroter_guidance_2021}. This new paradigm has been adopted by the European Commission and is encouraged in its technical topic guides for developing Sustainable Urban Mobility Plans (SUMP) \cite{fabian_kuster_practitioner_2019, jim_walker_supporting_2019}.

Therefore, the main interests for the purpose of this study are pedestrian spaces, road surfaces, and car parking surfaces.

\subsection{Semantic segmentation}

The most widely used deep learning models for image segmentation,
fully convolutional neural networks \cite{long_fully_2015},
employ image convolutions where the parameters of the kernel
are training parameters. Semantic segmentation models usually follow a 
encoder-decoder structure \cite{badrinarayanan_segnet_2016}. \\

Until recently fully convolutional neural networks where the standard for
semantic segmentation tasks. Recently language related tasks with deep
learning saw a large improvement with transformer based models
\cite{vaswani_attention_2023}. For image related tasks a vision transformer
was developed \cite{dosovitskiy_image_2020}

\begin{figure}[H]
    \centering
    \begin{tabular}{cc}
        \subcaptionbox{SAM model results when segmenting building roofs.}{\includegraphics[width=5cm,height=3cm]{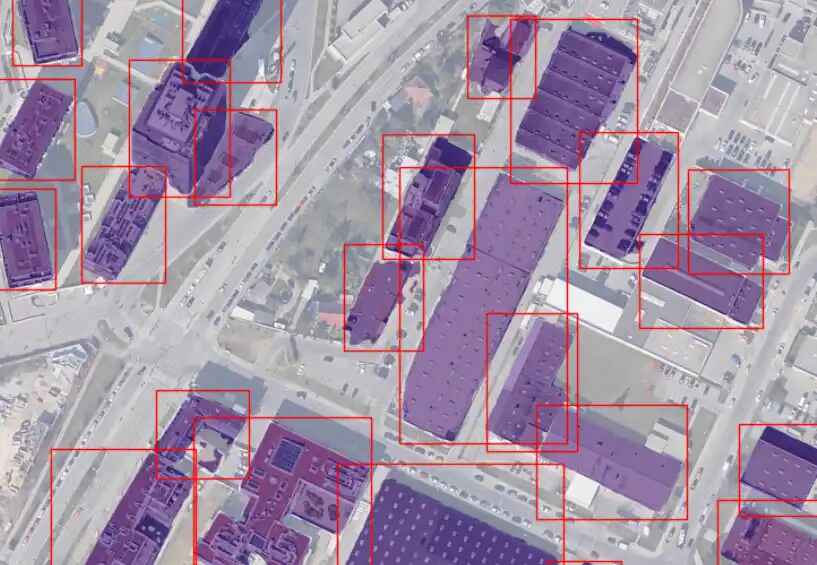}} &
        \subcaptionbox{SAM model results when segmenting parking spaces.}{\includegraphics[width=5cm,height=3cm]{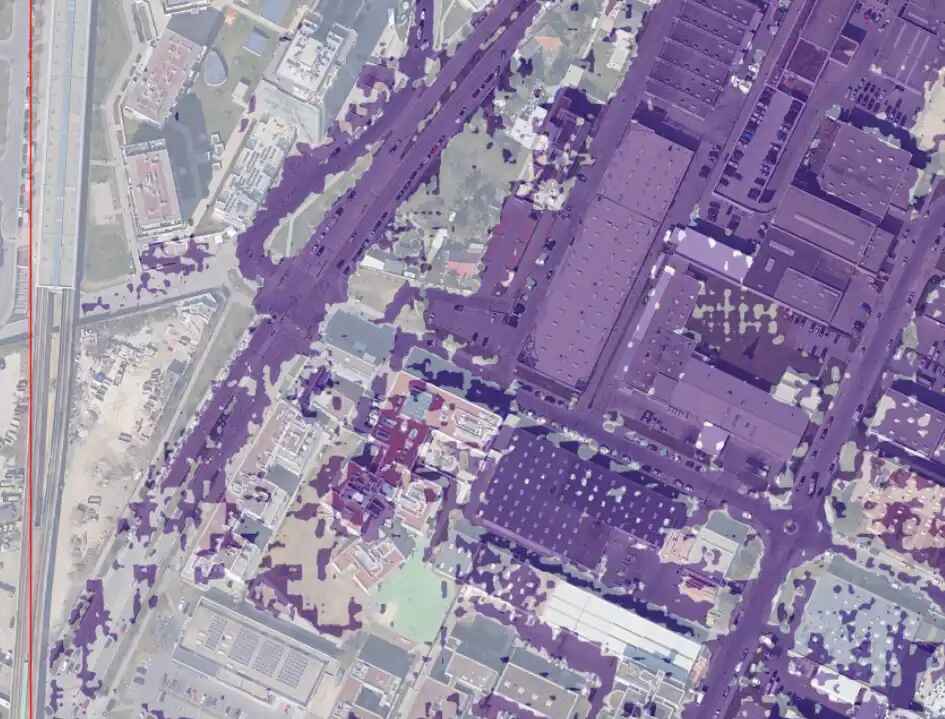}}
    \end{tabular}
    \caption{The SAM model is able to fulfil only more general and common tasks.}
    \label{fig:SAM}
\end{figure}

During the year 2023 two models revolutionized the research regarding image vision with deep learning. The DaTaSeg model \cite{gu_dataseg_2023} and the segment anything model SAM \cite{kirillov_segment_2023} both are one of the first general purpose semantic segmentation models. Those models are trained with very large datasets using
computing resources only available for the biggest technology companies. The SAM model includes a GPT module 
allowing any user to input an image and a text prompt. The model segments the image
according to the text given by the user (one shot segmentation). Another option provided by the 
model is to segment unknown objects in the image without providing any 
prompt or additional training (zero shot segmentation). 
A geospatial version 
SAM is available \cite{wu_samgeo_2023} through the 
SamGeo python package but employing the same trained model. The training dataset, the SA-1B dataset,  
comprises general images sourced from the internet \cite{kirillov_segment_2023} 
using information such as alt texts to train text promts, 
while the application involves aerial images, which possess markedly 
distinct characteristics and segmentation classes 
compared to the training data. 
General purpose semantic segmentation models such as 
SAM are effective in the context of geospatial 
analysis for general and easy tasks as segmenting buildings 
or detecting forests \cite{osco_segment_2023}, 
but for a specific
task like detecting parking spaces it fails completely [fig \ref{fig:SAM}]. \\

Transfer learning and fine tuning in the context of computer vision involves taking 
a pre-trained model, which has been trained on a general and diverse 
dataset and freezing the weights of the pre-trained 
model's encoder and train a new decoder. The underlying assumption 
is that a general understanding of image patterns and features is 
transferable, independent of the specific dataset chosen, no matter 
how different the datasets might be 
\cite{zhuang_comprehensive_2021}. However, it's 
important to note that this assumption has been empirically proven for years but the mathematical background is still under development \cite{cao_feasibility_2023}. \\

\section{Methodology}

\subsection{Machine learning implementation details}

\subsubsection{The model}

\begin{figure}[!htp]
    
    \centering
    \includegraphics[width=9cm]{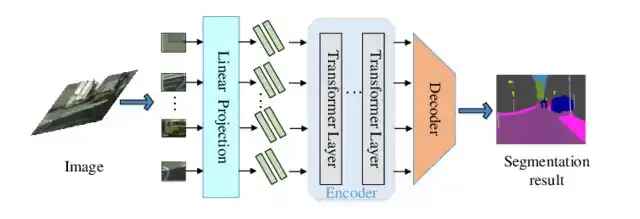}
    
    \caption{Transformer-based model \cite{dosovitskiy_image_2020}}
    \label{fig:transformer}
\end{figure}

For a deep learning approach an encoder-decoder structure is chosen. To achieve high quality results with limited resources, general purpose models such as SAM can be utilized using transfer learning 
for further training. 
The pre-trained encoder of the SAM model \cite{kirillov_segment_2023}
was employed, with only a new convolutional decoder being trained in
accordance with a U-Net like structure [fig \ref{fig:transformer}]. The pre-trained encoder from the SAM
model is a vision transformer trained with masked autoencoders on the
SA-1B dataset \cite{he_masked_2022,kirillov_segment_2023} 
for images with 1024x1024 pixel resultion. 
The implementation utilizes the pytorch-lightning \cite{falcon_pytorch_2023}
and segmentation models \cite{iakubovskii_segmentation_models_pytorch_2019}
Python libraries. For the initial results, the basic vision transformer encoder
with vit-b weights was utilized. \\

Loss functions for semantic segmentation tasks can be categorized into pixel,
boundary, and region-level loss functions \cite{azad_loss_2023}. In the case of training datasets featuring unbalanced classes, such as in this study, where smaller surfaces are detected over a main background class, the asymmetrical unified focal loss function \cite{yeung_unified_2022} is an adecuate loss function for model training. This combined loss  integrates the Tversky loss (region loss) with the focal loss (pixel loss) function. \\

The diversity of the images was measured calculating the dot product 
of the class count vector (number of pixels of each class in the image) of each image respect to the median value of the dataset. Images with lower diversity values are repeated more often during the training process.  

\subsubsection{Segmentation mask cleaning}

Segmentation results are cleaned using morphological operations from the \texttt{scikit-image} library \cite{van_der_walt_scikit-image_2014}. Holes in masks are closed, and noise is removed. Erosion thins objects by deleting border pixels, while dilation thickens objects by adding border pixels. Opening (erosion followed by dilation) removes noise, and closing (dilation followed by erosion) fills holes.

Erosion and dilation use image convolutions with specific kernels. Different kernels enhance various features \cite{jamil_noise_2008, serra_introduction_1986}. A round kernel rounds mask edges, a rectangular kernel creates rectangular shapes, and an octagon kernel sharpens edges optimally without limiting to rectangles \cite{srisha_morphological_2013}.

Segmentation classes have known area ranges, used to further clean results. For example, parking spaces should be at least $3 \text{ m}^2$ in area and $1.5 \text{ m}$ in width. Similar logic is applied to other classes.

The cleaning process is:

\begin{enumerate}
\item Erosion with a kernel of $\frac{1}{4}$ minimum width.
\item Deleting detections smaller than half the minimum area using binary opening.
\item Dilation.
\end{enumerate}

Then, the inverse:

\begin{enumerate}
\item Dilation.
\item Deleting dark spots smaller than $\frac{1}{4}$ minimum area using binary opening.
\item Erosion.
\end{enumerate}

This process is repeated for each class, treating other pixels as background.

\subsection{Aerial imagery datasets}

To illustrate the functionality of the developed code, two datasets were generated 
for the purpose of parking space detection in the cities of Vienna and Madrid. 
Creating the dataset in both scenarios can be accomplished with just a few 
lines of code, utilizing the library developed alongside this study. 
To generate the dataset, it is necessary to define a link to a WMTS 
service, specify the dataset area, and provide either a vectorial 
geometry file containing the ground truth or an overpass-turbo request. 

\subsubsection{Dataset grid}

\begin{figure}[!htp]
    \centering
    \begin{tabular}{cc}
        \begin{subfigure}[t]{0.47\linewidth}
            \centering
            \includegraphics[width=6cm, height=3.5cm]{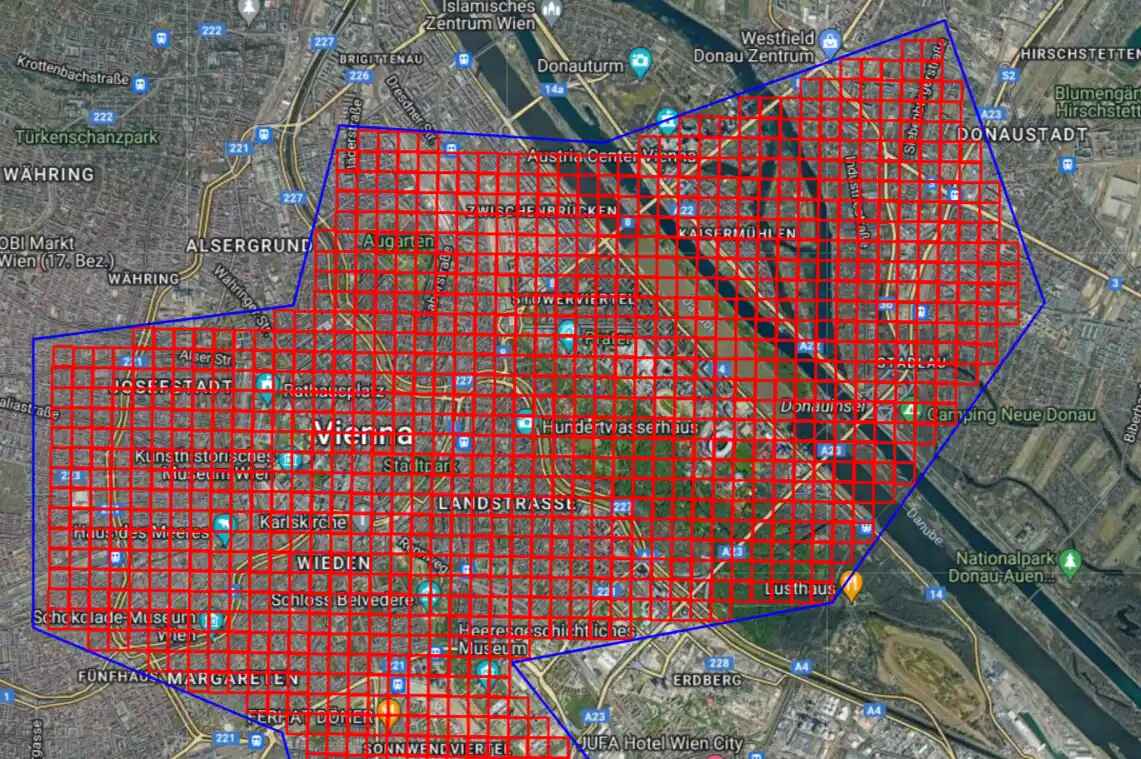}
            \caption{Dataset region (blue) and generated dataset grid (red).}
            \label{fig:grid_exaple}
        \end{subfigure} &
        \begin{subfigure}[t]{0.47\linewidth}
            \centering
            \includegraphics[width=6cm, height=3.5cm]{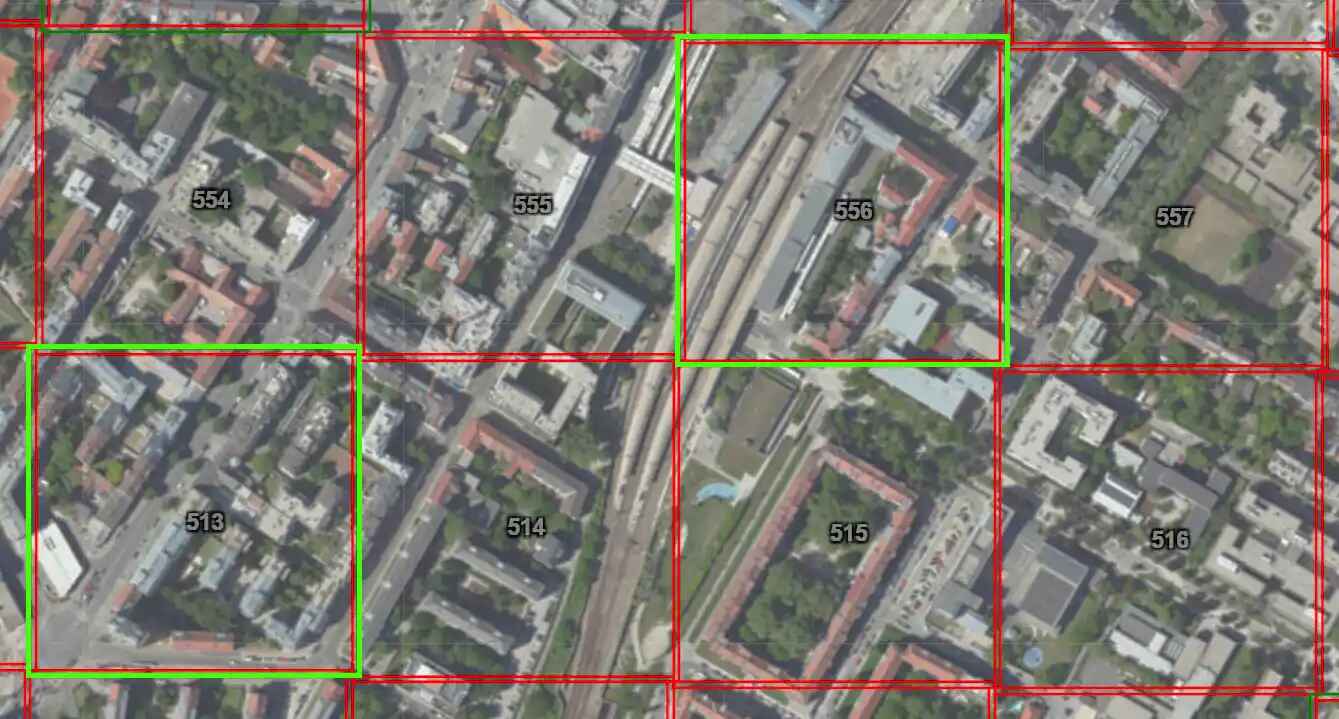}
            \caption{Detail of the tiles in the grid. In green tiles randomly selected for the dataset. Numbers are the tile ids. Overlap is set to 0 m.}
            \label{fig:grid_detail}
        \end{subfigure}
    \end{tabular}
    \caption{Dataset grid example.}
    \label{fig:grid}
\end{figure}

Neural networks based on Transformers need input images with a particular pixel
size, whereas convolutional networks can handle images of different sizes. It is
recommended for all images in the dataset to have similar sizes and resolutions in
both scenarios, as these aspects can influence the visual characteristics of specific
features in the images. To form a dataset, one or more polygons are assigned as the
dataset region, and a grid structure is created [fig \ref{fig:grid}].

The procedure involves defining a grid in UTM coordinates, shaped as a rectangle,
based on the overall bounds of the dataset region. This grid is established with
a specified tile size and overlap between tiles. As minimum and maximum values for x and y coordinates are needed to georeference a matrix and save it as a image, tiles are defined as minimum and maximum x and y values. After forming the grid, all coordinates are
converted to the coordinate system of the image provider. This conversion might
adjust the grid bounds, as they could appear rotated. The reason for this is that the x and y axis can rotate during the coordinate conversion and, to cover the minimum and maximum x and y values from the previous coordinate system a larger area has to be selected and tiles can overlap slightly. In figure \ref{fig:grid} tiles exhibit this situation due to the aerial image being in geographic coordinates
while the grid is in UTM resulting in a small misalignment and overlap of the tile bounds. The grid is then converted to the coordinates of the
image to accurately preserve the image bounds. Similarly, the mask in vector
format undergoes conversion to the image's coordinate system. Incorporating
overlap between tiles could be important for precise results at the image borders,
helping prevent errors caused by partially visible objects. The overlap can
also be configured accordingly. 

\subsubsection{Aerial imagery}
Aerial optical imagery is globally accessible. Satellites provide lower-resolution images, with NASA's Landsat 8 and 9 offering 15-meter resolution and an 8-day revisit time \cite{noauthor_usgs_nodate}, and ESA's Sentinel 2 providing 10-meter resolution with a 5-day revisit time \cite{noauthor_copernicus_nodate}. Both agencies offer WMTS services for true-color images covering the entire Earth. Private companies like DigitalGlobe provide higher-resolution imagery (up to 30 cm), but access is limited and requires payment \cite{digitalglobe_worldview-3_nodate}.

Drone or airplane images offer higher resolutions, typically 20 to 5 cm per pixel. For semantic segmentation and urban feature detection, resolutions finer than 1 meter are often needed [fig \ref{fig:image_res
}]. Our parking space detection project required images better than 20 cm per pixel, which can be challenging in regions with only satellite imagery.

\begin{figure}[!htp]
\centering
\begin{tabular}{cc}
\begin{subfigure}[t]{0.45\linewidth}
\centering
\includegraphics[width=4.5cm, height=3cm]{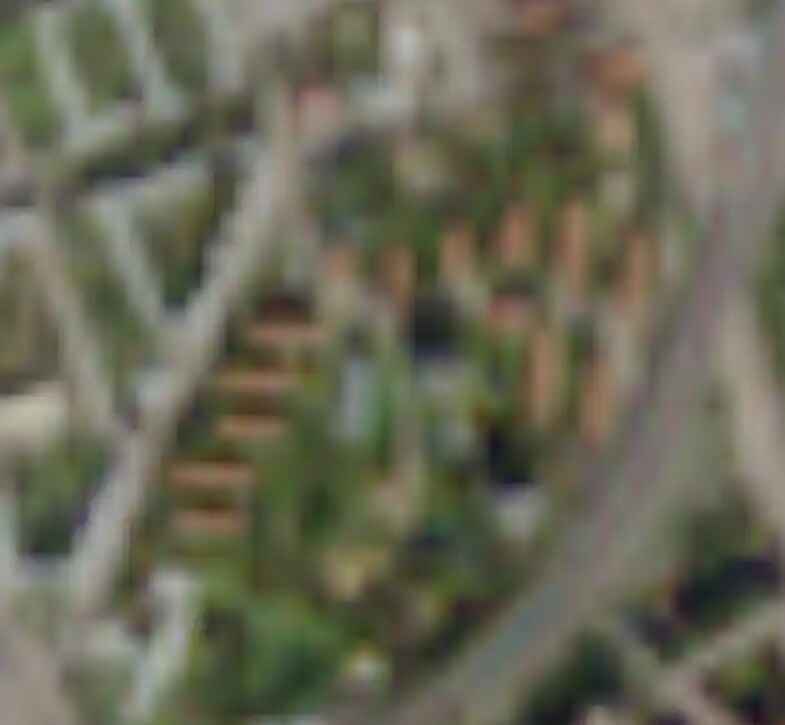}
\caption{Sentinel 2 10 m resolution image in Vienna \cite{noauthor_copernicus_nodate}.}
\label{fig:vienna_10m
}
\end{subfigure} &
\begin{subfigure}[t]{0.45\linewidth}
\centering
\includegraphics[width=4.5cm, height=3cm]{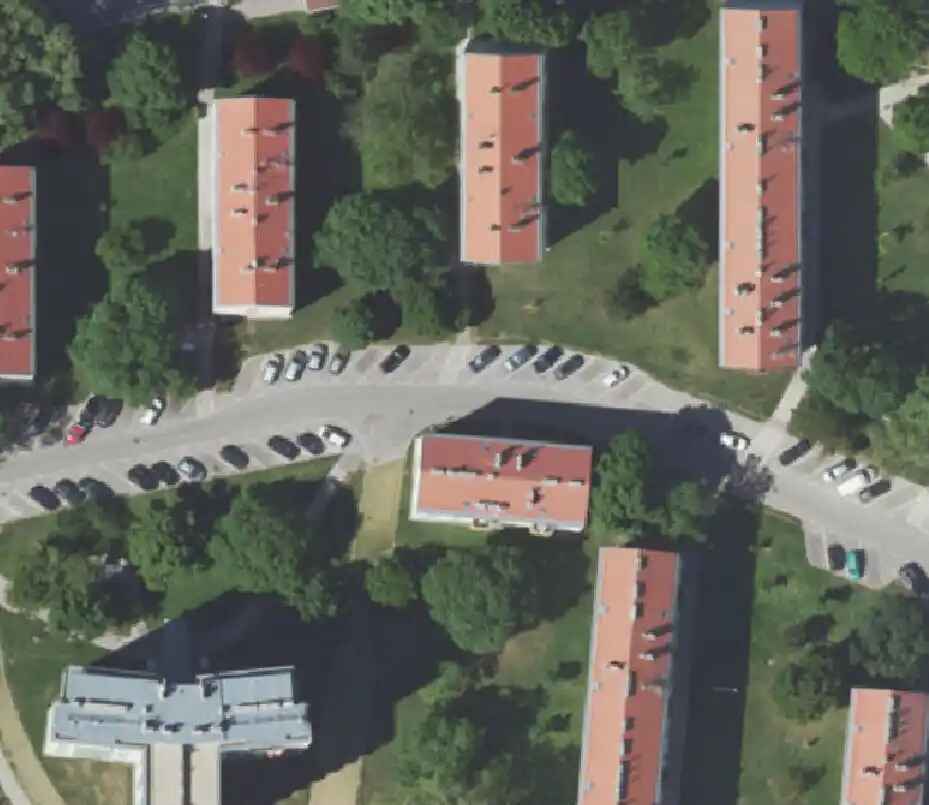}
\caption{20 cm resolution orthophoto in the same area \cite{noauthor_stadtvermessung_nodate}.}
\label{fig:vienna_20_cm
}
\end{subfigure}
\end{tabular}
\caption{The importance of image resolution.}
\label{fig:image_res
}
\end{figure}

Google Earth \cite{noauthor_google_nodate} offers global aerial photography with resolutions from 15 meters to 10 cm. However, controlling the date and quality of images is challenging due to stitching from multiple sources. 

European cities provide public vector cartography and orthophotography with resolutions of 20 to 5 cm, updated annually \cite{noauthor_city_nodate, noauthor_stadtvermessung_nodate, paris_city_council_paris_nodate, noauthor_luftbilder_nodate, ayuntamiento_de_madrid_geoportal_nodate}. These datasets are available via cloud services following Open Geospatial Consortium standards \cite{maidment_open_2011}. Historic aerial images since 1950 are often available. Overall, sufficient resolution imagery is available in most regions.

Aerial images have distortions from camera tilt or terrain topography. Orthorectification removes these distortions, allowing precise measurements. This is crucial in urban environments, as buildings can occlude streets and distort building footprints [fig \ref{fig:distortion
}].

\begin{figure}[!htp]
\centering
\begin{tabular}{ccc}
\begin{subfigure}[t]{0.3\linewidth}
\centering
\includegraphics[width=4cm, height=3cm]{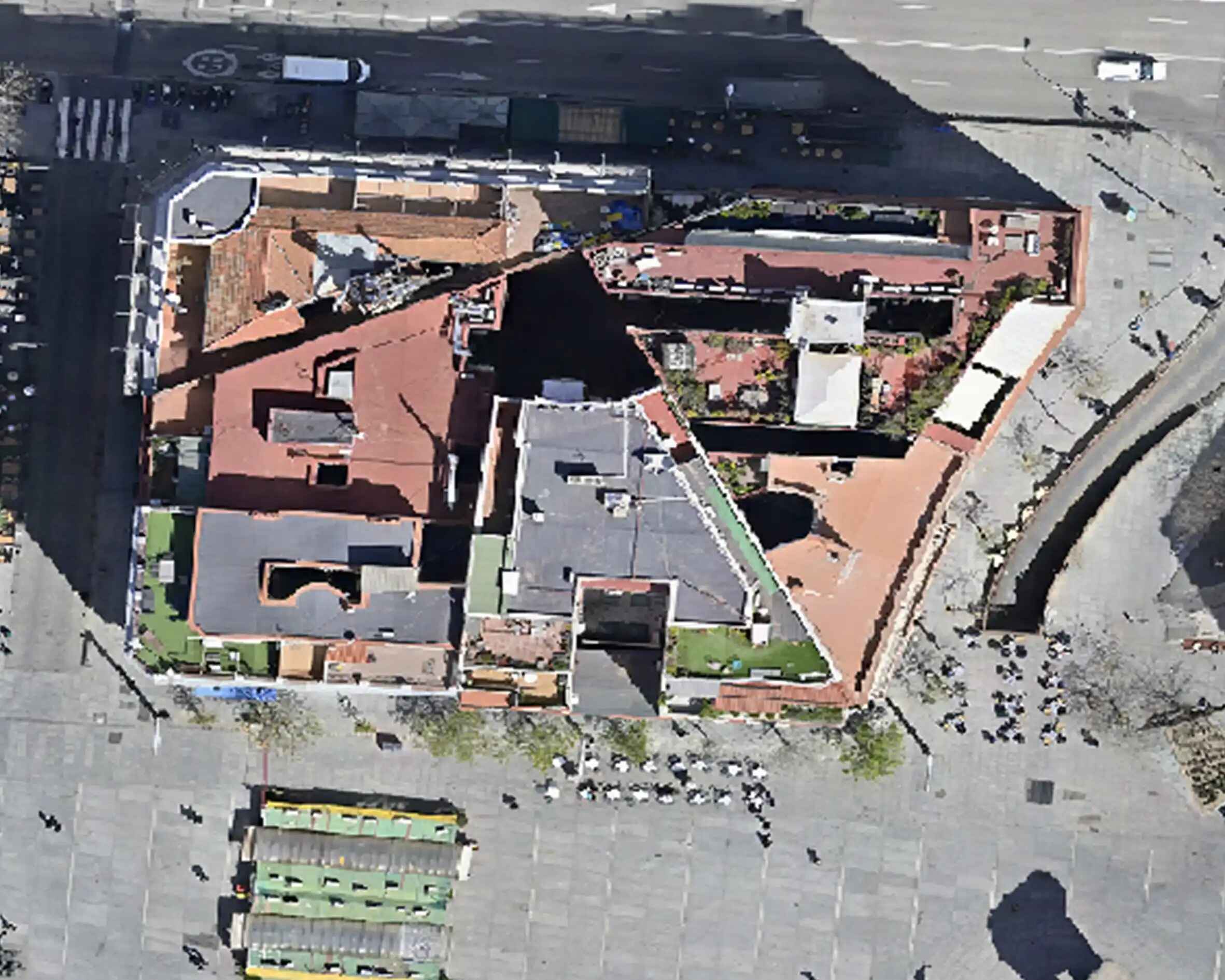}
\caption{1.5 cm/pix UAV image.}
\label{fig:1.5cma
}
\end{subfigure}
&
\begin{subfigure}[t]{0.3\linewidth}
\centering
\includegraphics[width=4cm, height=3cm]{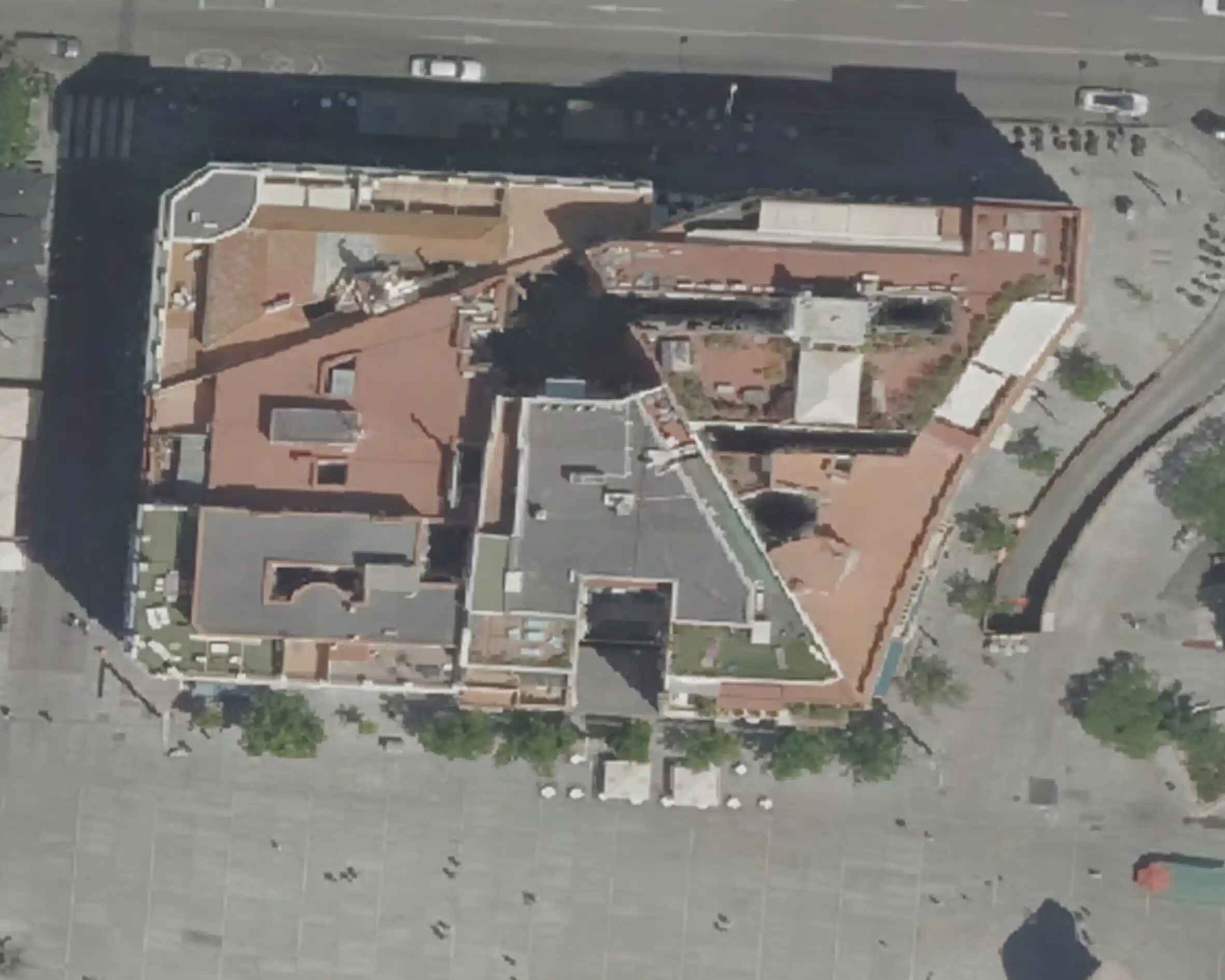}
\caption{10 cm/pix Aerial image.}
\label{fig:10cmb
}
\end{subfigure}
&
\begin{subfigure}[t]{0.3\linewidth}
\centering
\includegraphics[width=4cm, height=3cm]{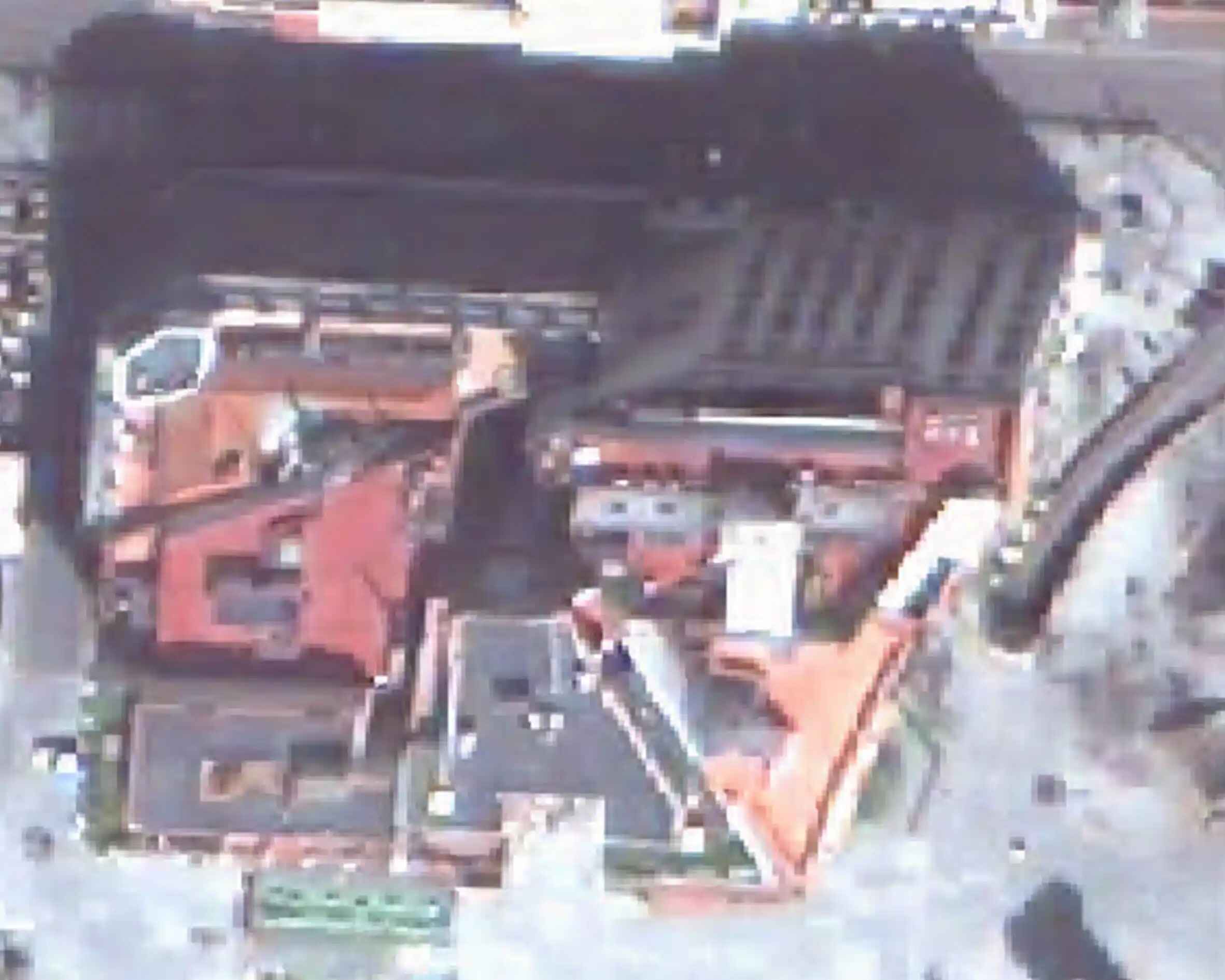}
\caption{30 cm/pix Satellite (WorldView3) image.}
\label{fig:30cmc
}
\end{subfigure}
\end{tabular}
\caption{Remaining distortion in orthoimages of a building in Madrid \cite{ayuntamiento_de_madrid_geoportal_nodate} taken by UAV (least distortion), plane, and satellite (most distortion).}
\label{fig:distortion
}
\end{figure}

Image resolution and orthorectification are vital for classes like road surface, parking spaces, or sidewalks. Objects near buildings may be occluded, and trees pose challenges. Winter images have larger shadows, while summer images have less shadows but may obscure streets with trees [fig \ref{fig:issues_providers
}].

\begin{figure}[!htp]
\centering
\begin{tabular}{cc}
\begin{subfigure}[t]{0.4\linewidth}
\centering
\includegraphics[width=4.5cm, height=3cm, ]{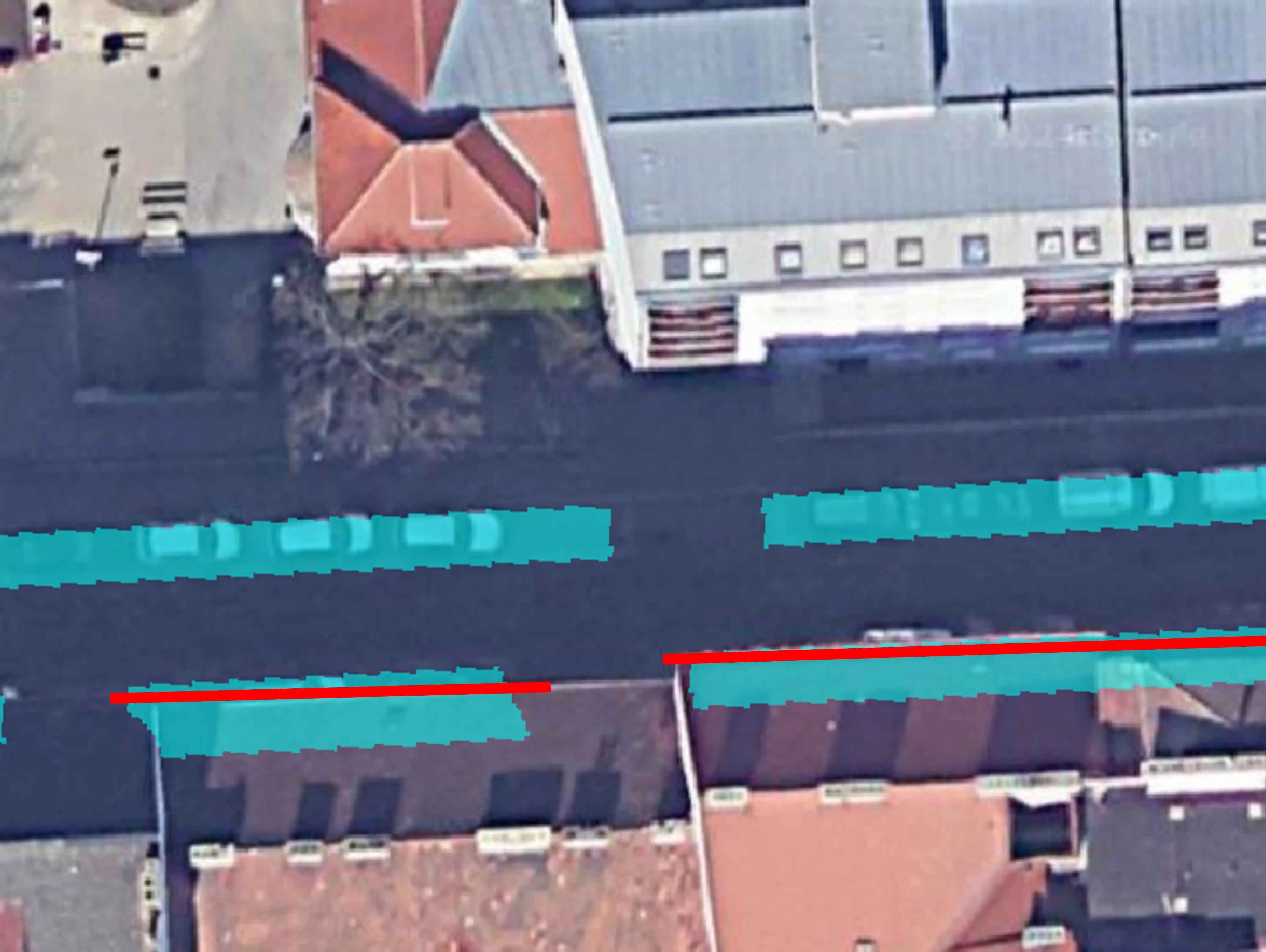}
\caption{Vienna winter image 2021 \cite{noauthor_google_nodate}.}
\label{fig:vienna_winter_a
}
\end{subfigure} &
\begin{subfigure}[t]{0.4\linewidth}
\centering
\includegraphics[width=4.5cm, height=3cm, ]{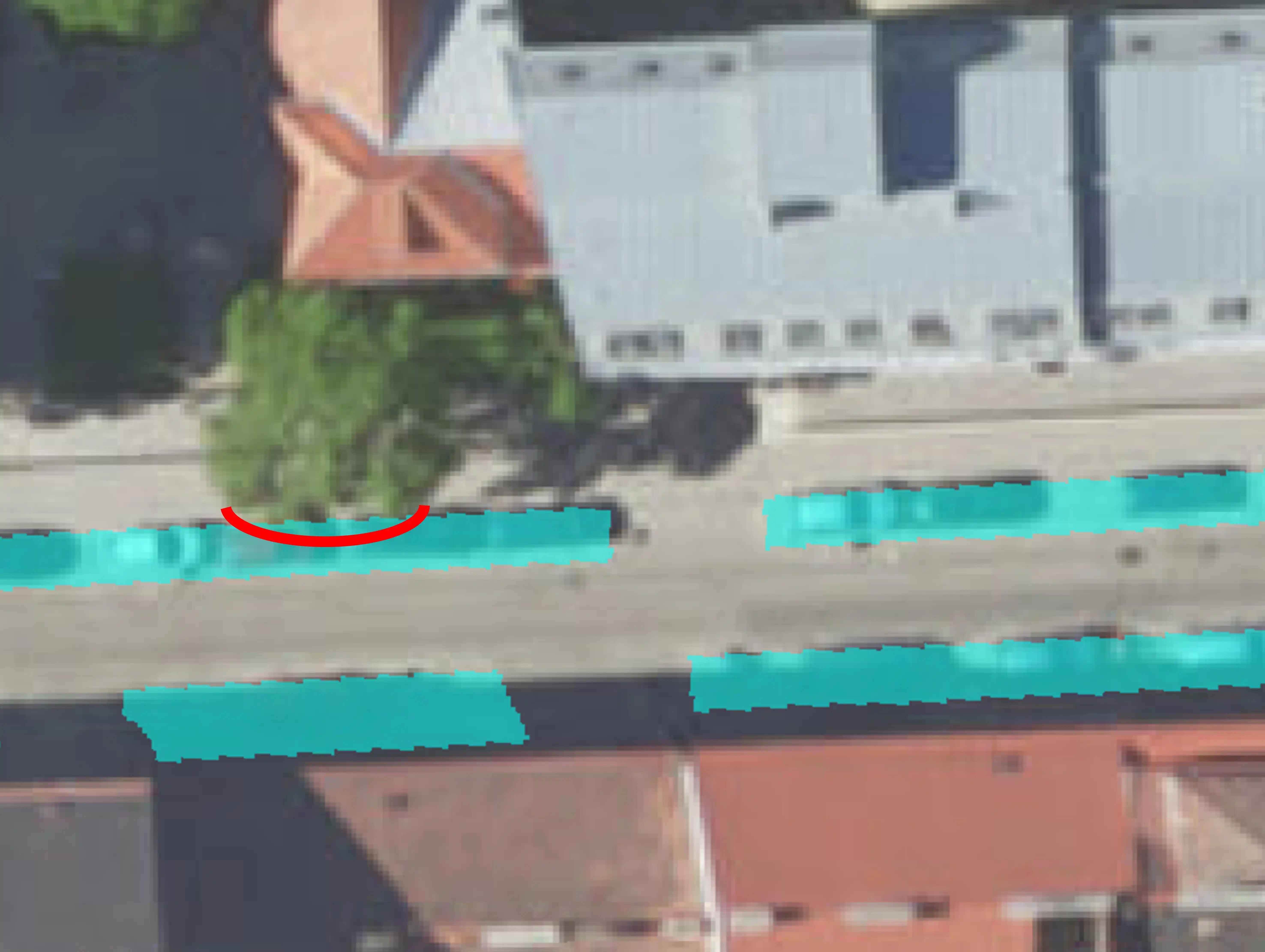}
\caption{Vienna summer image 2022 \cite{noauthor_stadtvermessung_nodate}.}
\label{fig:vienna_summer_b
}
\end{subfigure}
\end{tabular}
\caption{Issues with different image providers.}
\label{fig:issues_providers
}
\end{figure}

\subsubsection{Ground truth}

Ground truth data must be in the form of vector data, represented as polygons that delineate the objects targeted for segmentation. The use of vector data allows for seamless compatibility with various coordinate systems and resolutions. It is crucial to maintain the coordinate system used by the image provider consistently, necessitating the conversion of both the grid and ground truth to the coordinate system of the images. If the ground truth is initially presented as raster data, it can be transformed into vector data using tools such as rasterio \cite{gillies_rasterio_2013}. \\

A challenge arises when cartography is available as line data instead of polygons. 
Consequently, polygons need to be generated by connecting the lines that enclose 
the desired objects. This adds inaccuracies in the data. 

The main inaccuracies found on the data used for this project are: 

\begin{itemize}
    \item Different or unclear criteria from the responsible organization when classifying the data. 
    \item Even though the organization states the data where updated, new infrastructure or changes shown on the images are missing. 
    \item Mistakes or inaccuracies in the boundaries of object.
    \item Inaccuracies or missing data when converting from line to polygon geometry. 
\end{itemize}

\subsubsection{Madrid dataset}

Two version of the dataset were created [figure \ref{fig:MadGrid}]. The main version has 6 classes: background, building, road, sidewalk, swimming pool, bike path and parking. The swimming pool class was chosen as it is considered an important factor for suburbanization and car-centric developments in Spain. The second version only has parking spaces. The selection of the training and testing areas aimed for diversity, encompassing poorer neighborhoods in the southern periphery characterized by dense and unorganized urban structures, newer neighborhoods from the 2000s in the eastern region featuring grid street patterns and multi-family housing with shared swimming pools, richer areas in the north with high rise buildings from the 80s and 90s, as well as neighborhoods from the historic medieval center and 19th-century developments.

The train dataset (4778 images and masks with a size of 1024x1024 pixels) encompasses all tiles contained inside the official boundaries of the following neighbourhoods of Madrid: 
Berruguete, Costillares, El Viso,Castellana, Quintana, Embajadores, Puerta del Angel, Los Rosales, Acacias, Goya, Numancia, Palomeras Bajas, Valderrivas, Orcasitas, Almendrales, Universidad, Almagro, Bellas Vistas and Hispanoamerica.

The test dataset (1238 images and masks with a size of 1024x1024 pixels) encompasses all tiles contained inside the official boundaries of the following neighbourhoods of Madrid. It is used exclusively for evaluation: Gaztambide, Cuatro Caminos, Pilar, Arcos and San Diego.

\begin{figure}[!htp]
    \centering
    \begin{tabular}{ccc}
        \begin{subfigure}[t]{0.31\linewidth}
            \centering
            \includegraphics[width=4cm, height=3cm]{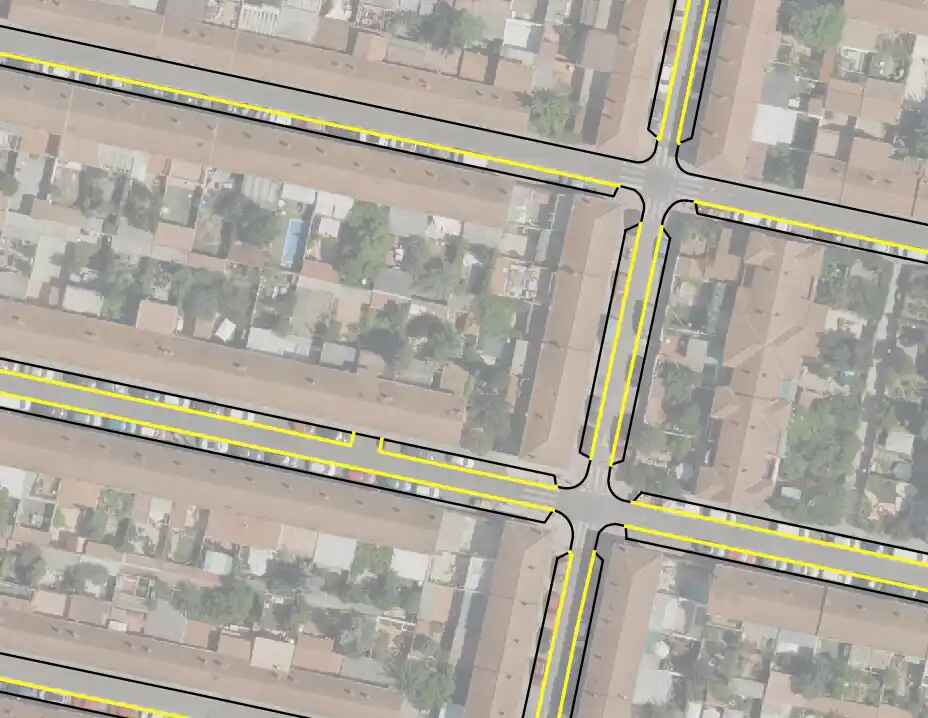}
            \caption{Madrid's cartography provides parking spaces as lines (yellow) \cite{ayuntamiento_de_madrid_geoportal_nodate}.}
            \label{fig:parkinglines}
        \end{subfigure} &
        \begin{subfigure}[t]{0.31\linewidth}
            \centering
            \includegraphics[width=4cm, height=3cm]{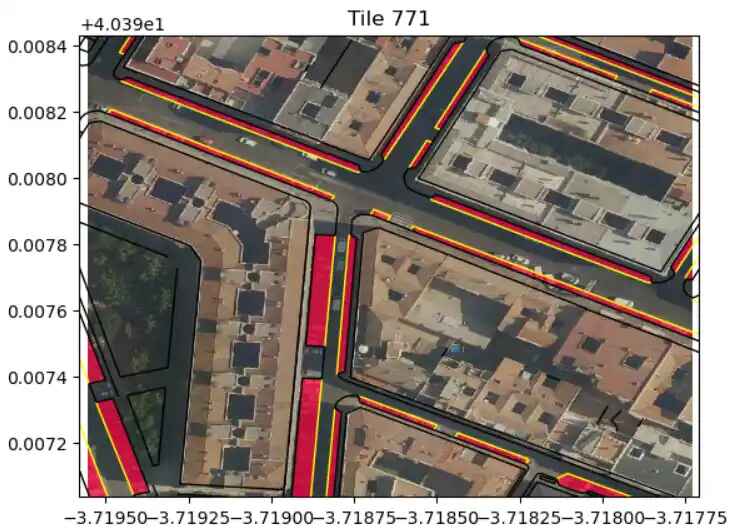}
            \caption{Tile 771 of the dataset. In red the parking polygons in raster format created using the available line geometry.}
            \label{fig:madtile771}
        \end{subfigure} &
        \begin{subfigure}[t]{0.31\linewidth}
            \centering
            \includegraphics[width=4cm, height=3cm]{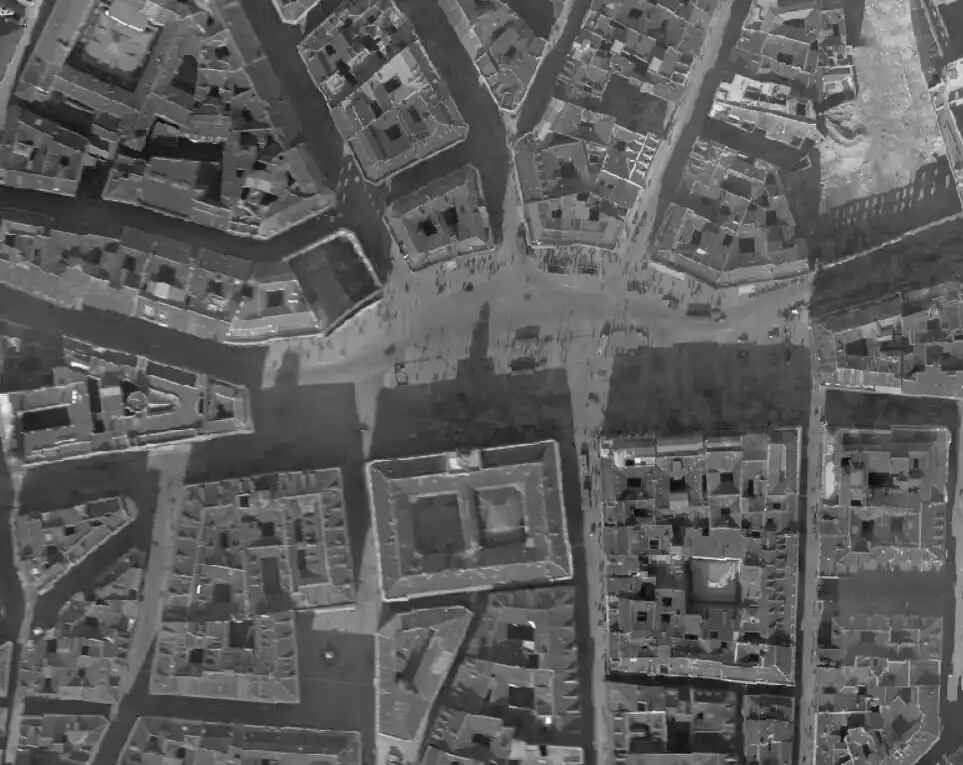}
            \caption{High resolution historic image from 1941 \cite{ayuntamiento_de_madrid_geoportal_nodate}.}
            \label{fig:mad1941}
        \end{subfigure}
    \end{tabular}
    \caption{Madrid dataset details.}
    \label{fig:madrid_ds_exaple}
\end{figure}

\begin{figure}[!htb]
    \centering
    \begin{tabular}{cccc}
        \begin{subfigure}[t]{0.24\linewidth}
            \centering
            \includegraphics[width=\linewidth, height=2.5cm]{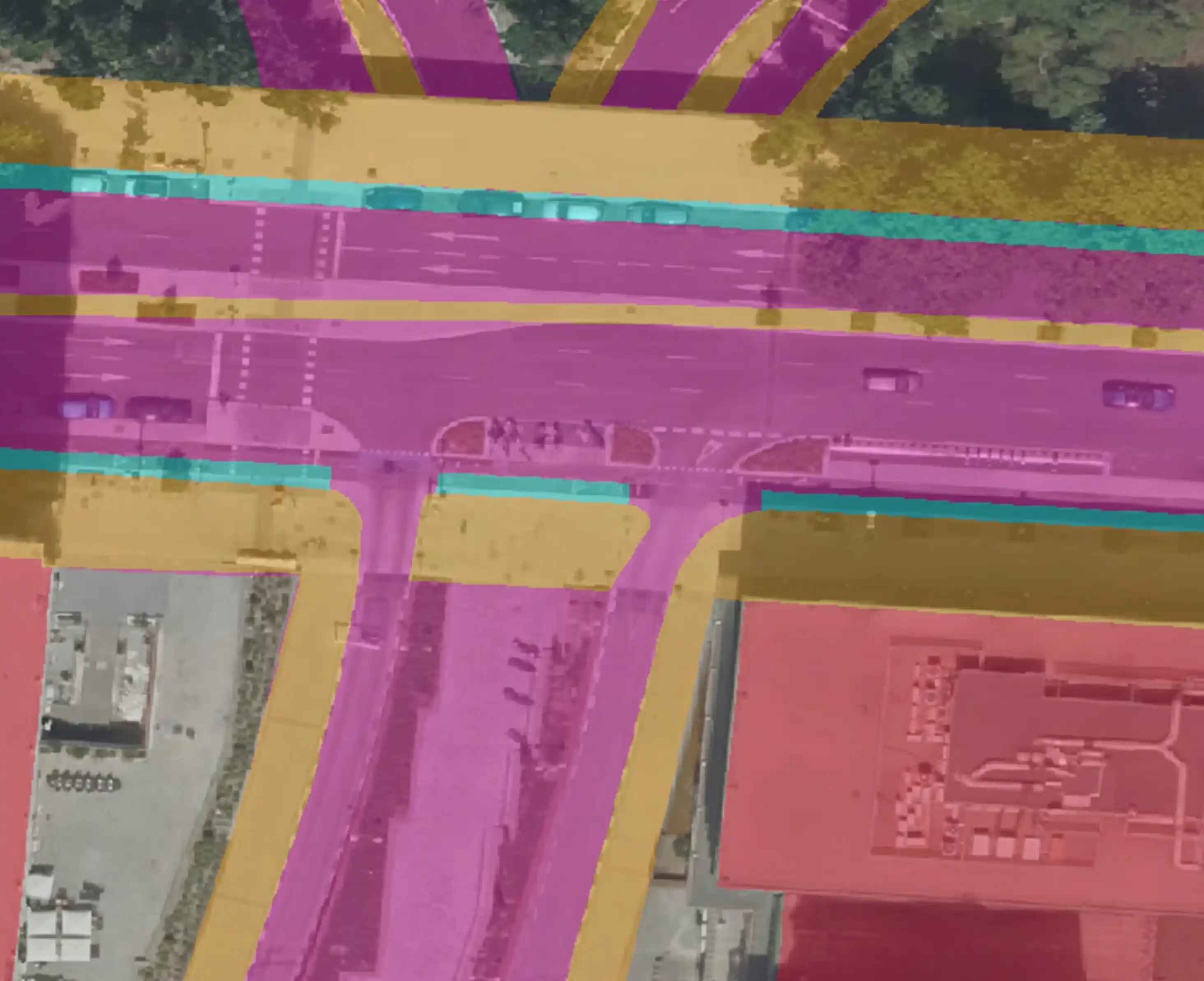}
            \caption{Ground truth data in the test set is missing the new bike lane. Parking and sidewalk were displaced too.}
            \label{fig:wrong_gt_1_mad}
        \end{subfigure} &
        \begin{subfigure}[t]{0.24\linewidth}
            \centering
            \includegraphics[width=\linewidth, height=2.5cm]{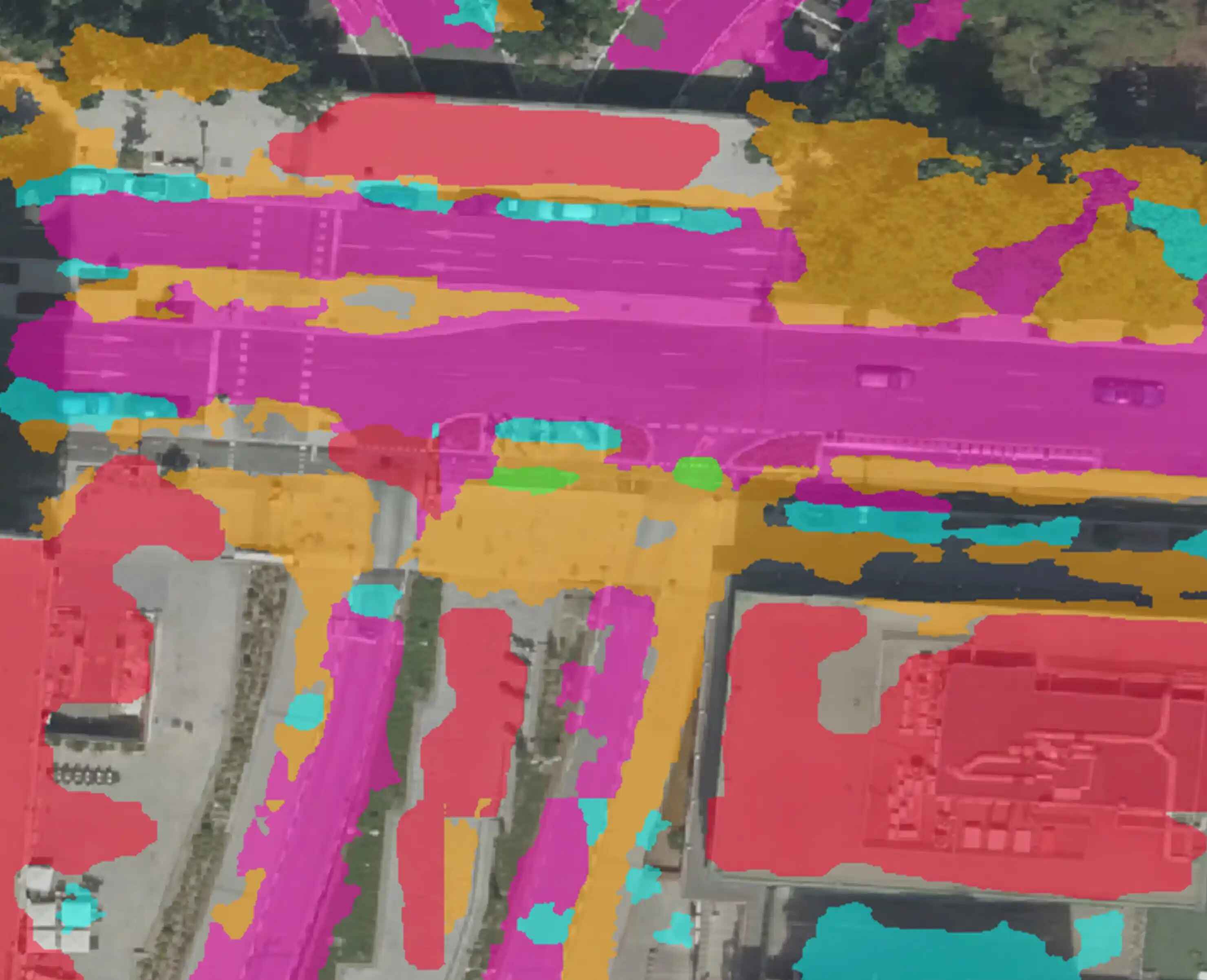}
            \caption{Model's results show inaccurate results. Some bike lanes are detected, but most are missing.}
            \label{fig:wrong_1_mad}
        \end{subfigure} &
        \begin{subfigure}[t]{0.24\linewidth}
            \centering
            \includegraphics[width=\linewidth, height=2.5cm]{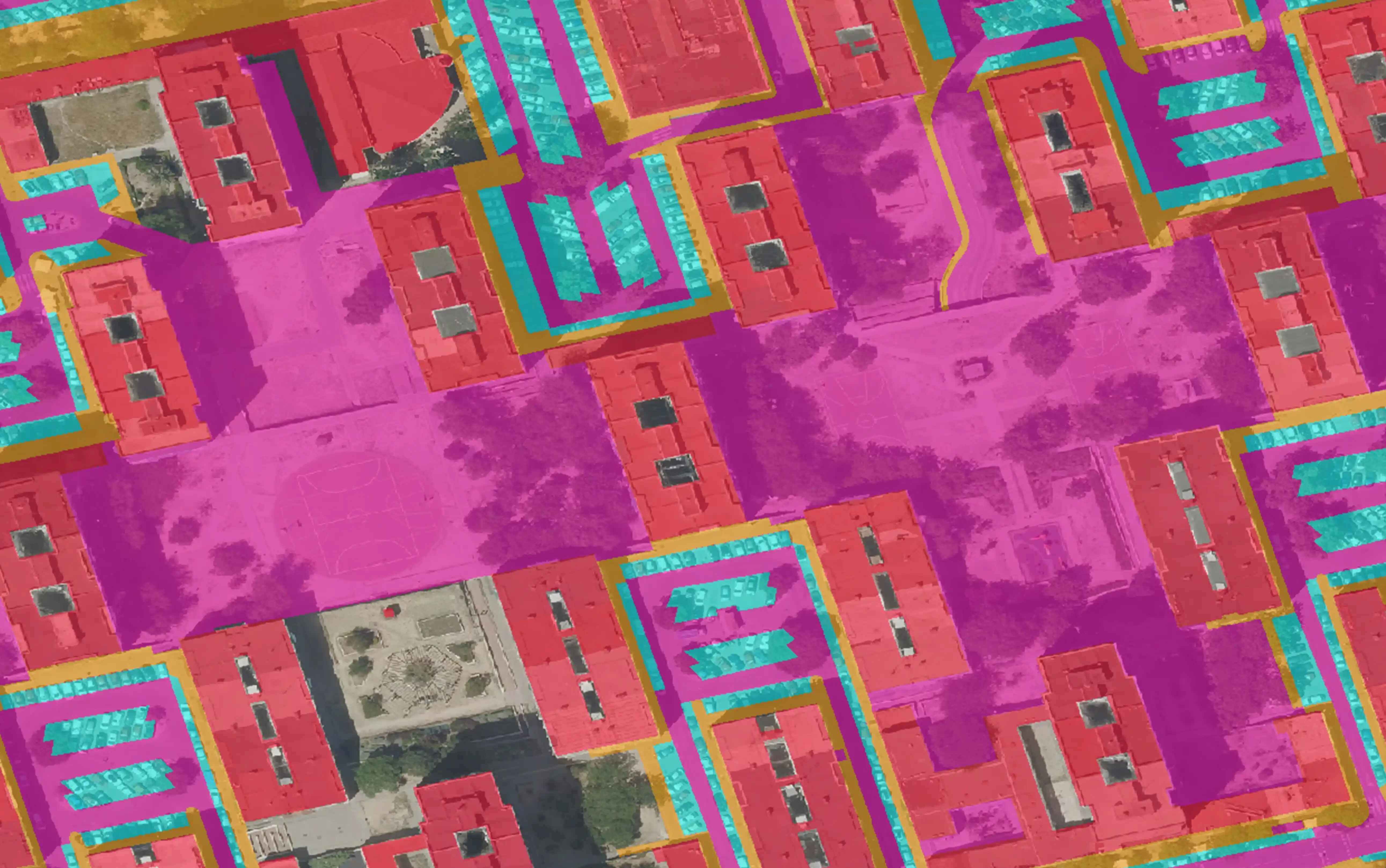}
            \caption{Ground truth shows big areas classified as road that are wrong.}
            \label{fig:wrong_gt_2_mad}
        \end{subfigure} &
        \begin{subfigure}[t]{0.24\linewidth}
            \centering
            \includegraphics[width=\linewidth, height=2.5cm]{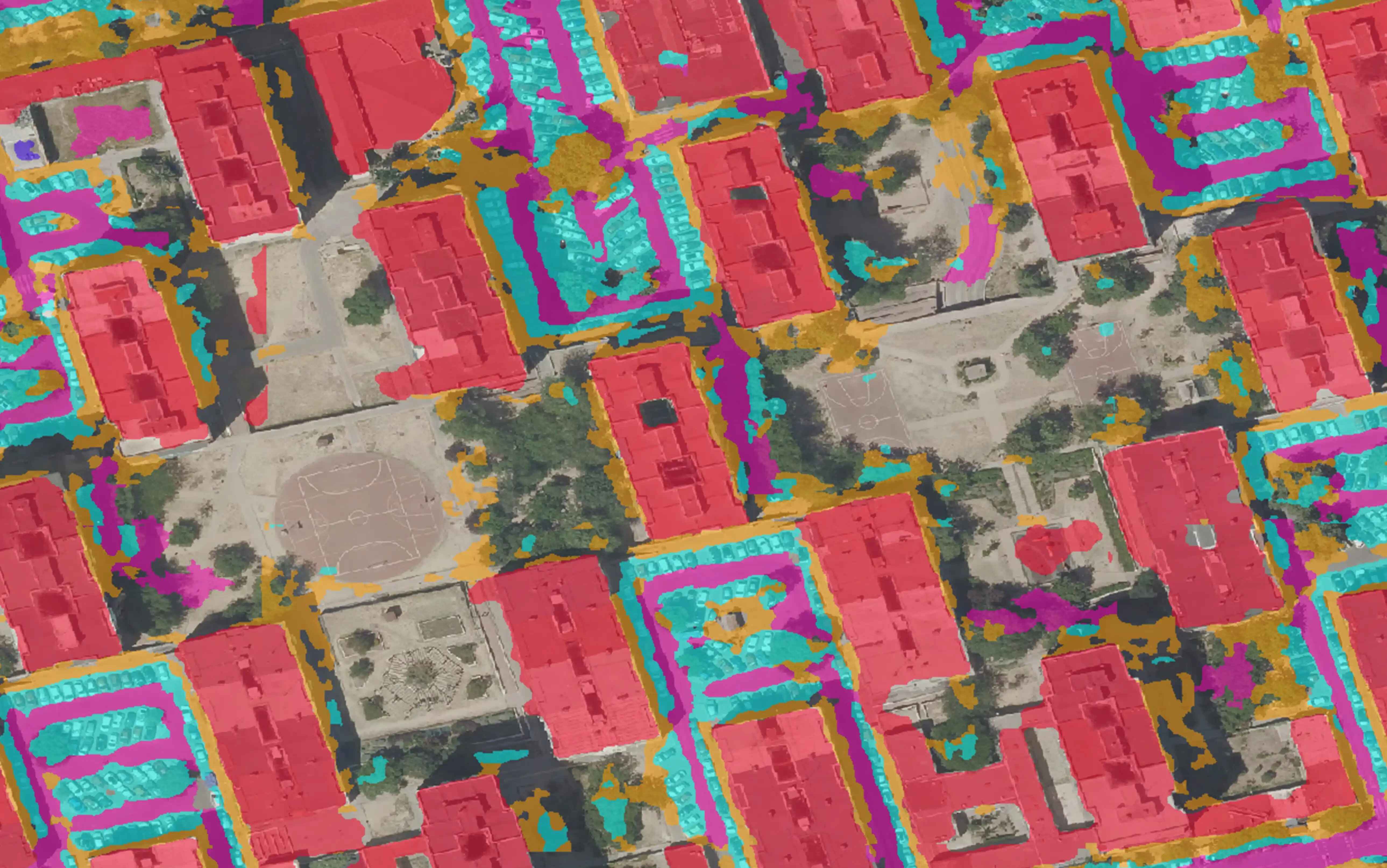}
            \caption{The model correctly does not classify the wrong areas as road but this will be negatively evaluated.}
            \label{fig:wrong_2_mad}
        \end{subfigure}
    \end{tabular}
    \caption{Some examples about mistakes in the ground truth from the Madrid test set.}
    \label{fig:madwrong}
\end{figure}

For the conversion from line to polygon geometry for the parking class the accuracy of the data was measured calculating the length of the parking lines contained in the parking polygon boundaries. The ratio over the total length of the parking lines show that the worst neighbourhood passes 85\% and that usual values where over 90\%.

The inaccuracies found in the Madrid dataset are especially related to unclear criteria to classify data and data that is not up to date even though the 2023 cartography was used. The figure \ref{fig:madwrong} shows some examples of the problems found in the dataset and the models output.

\subsubsection{Vienna Dataset}

\begin{figure}[!htb]
    \centering
    \begin{tabular}{cc}
        \begin{subfigure}[t]{0.47\linewidth}
            \centering
            \includegraphics[width=5cm, height=4cm]{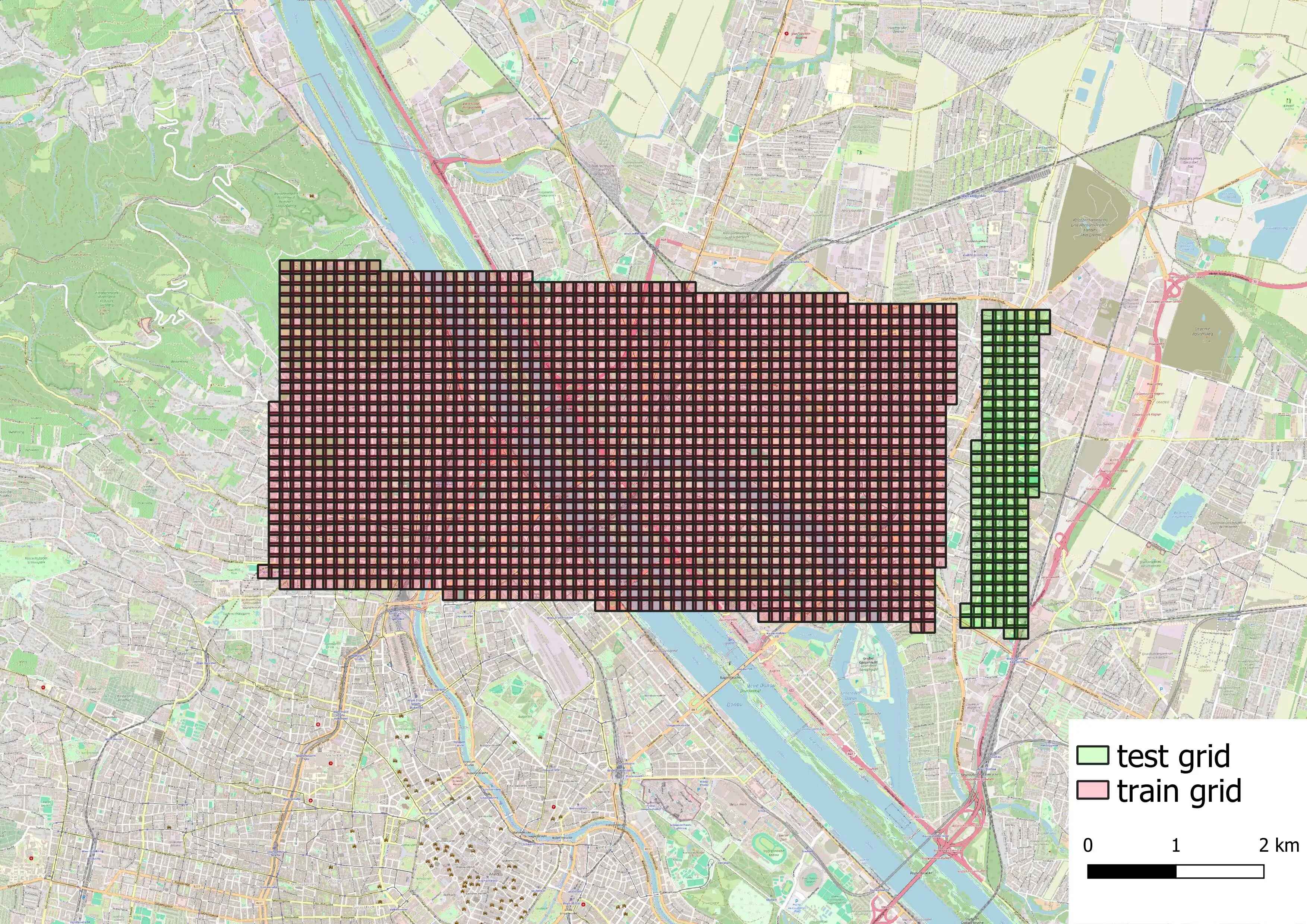}
            \caption{Training and testing areas.}
            \label{fig:VieGrid}
        \end{subfigure} &
        \begin{subfigure}[t]{0.47\linewidth}
            \centering
            \includegraphics[width=5cm, height=4cm]{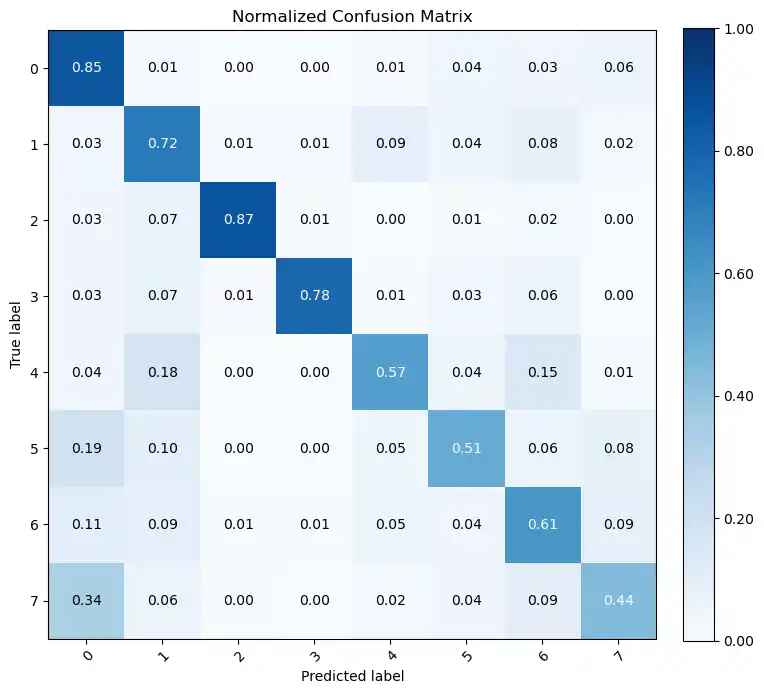}
            \caption{Normalized confusion matrix for the Vienna test set (Class 0 is background).}
            \label{fig:CMVienna}
        \end{subfigure}
    \end{tabular}
    \caption{Vienna dataset and model results.}
    \label{fig:resVie}
\end{figure}

The Vienna dataset encompasses a rectangular region situated to the east and west 
of the Danube, combining old and dense urbanisation with newer developments and even single family homes [fig \ref{fig:VieGrid}].  

Two versions of the Vienna Dataset were created. The main version has 7 categories: background, public road, tram or train tracks, crosswalk, on street parking, private road surface, sidewalk and separated pedestrian or bicycle path. The dataset has 1826 images for training and 156 for testing all with a size of 1024x1024 pixels. Another version with 2 classes (background and on street parking) was created only to detect parking spaces. 

\begin{figure}[!htb]
    \centering
    \begin{tabular}{cc}
        \begin{subfigure}[t]{0.47\linewidth}
            \centering
            \includegraphics[width=5cm, height=3cm]{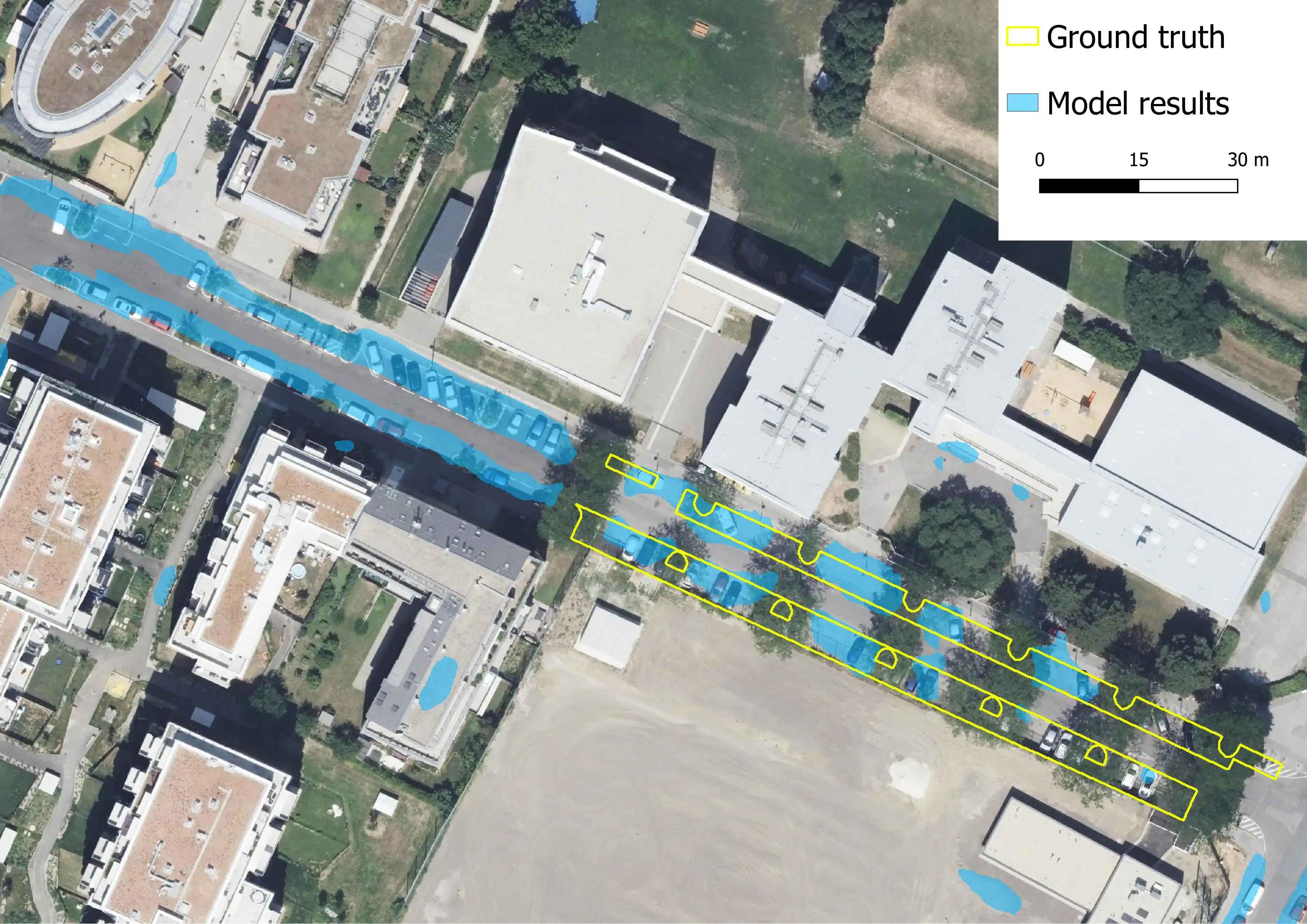}
            \caption{Ground truth data in the test set is missing some parking spaces. The model's results correctly classify all parking spaces as parking, but it will be negatively evaluated as most of the parking is missing in the ground truth.}
            \label{fig:wrong_vie_1}
        \end{subfigure} &
        \begin{subfigure}[t]{0.47\linewidth}
            \centering
            \includegraphics[width=5cm, height=3cm]{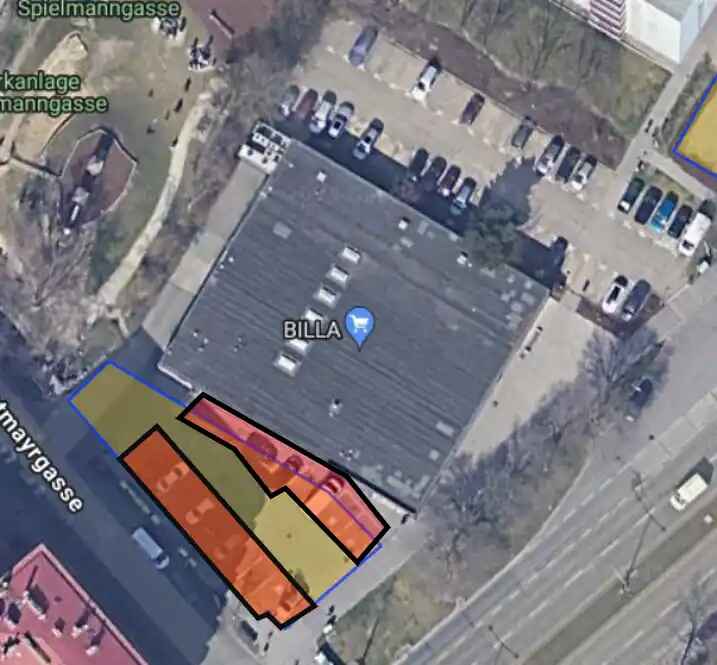}
            \caption{Overpass-turbo request results (in yellow) are not accurate enough compared to actual parking spaces (in red), and the data is incomplete. (In the top right corner, there is an unlabeled parking area)}
            \label{fig:wrong_vie_osm}
        \end{subfigure}
    \end{tabular}
    \caption{Examples of mistakes in the ground truth from the Vienna test set.}
    \label{fig:viewrong}
\end{figure}

Especially the OSM requests have many mistakes, as shown in figure \ref{fig:wrong_vie_osm}. In the case of the Vienna dataset, most issues are related to missing data, particularly for the parking class, both in OSM and in the government's data [fig \ref{fig:viewrong}].

\section{Results}

The metrics chosen for model evaluation are: 

\begin{itemize}
    \item \textbf{IoU}: Jaccard Index or Intersection over Union (IoU). 
    \item \textbf{F1 score}: F1 score or Dice score.
    \item \textbf{IoU\_200}: IoU using a 200-centimeter buffer for the ground truth before calculating the intersection.
    \item \textbf{Street ratio}: Amount of the model's output falling inside one of the classes related to streets in ground truth data (classes 2 and 6).
    \item \textbf{Pedestrian ratio}: Amount of the model's output falling inside one of the classes related to pedestrians in ground truth data (classes 3 and 5).
    \item \textbf{$\frac{\text{model area}}{\text{GT area}}$}: Area of the model output with class \(i\) divided by the area of the ground truth with class \(i\). This ratio measures the amount of over or under detection in the model.
\end{itemize}

\subsection{Madrid}

\subsubsection{Model evaluation}

\begin{figure}[!htb]
    \centering
    \begin{tabular}{cc}
        \begin{subfigure}[t]{0.47\linewidth}
            \centering
            \includegraphics[width=5cm, height=4cm]{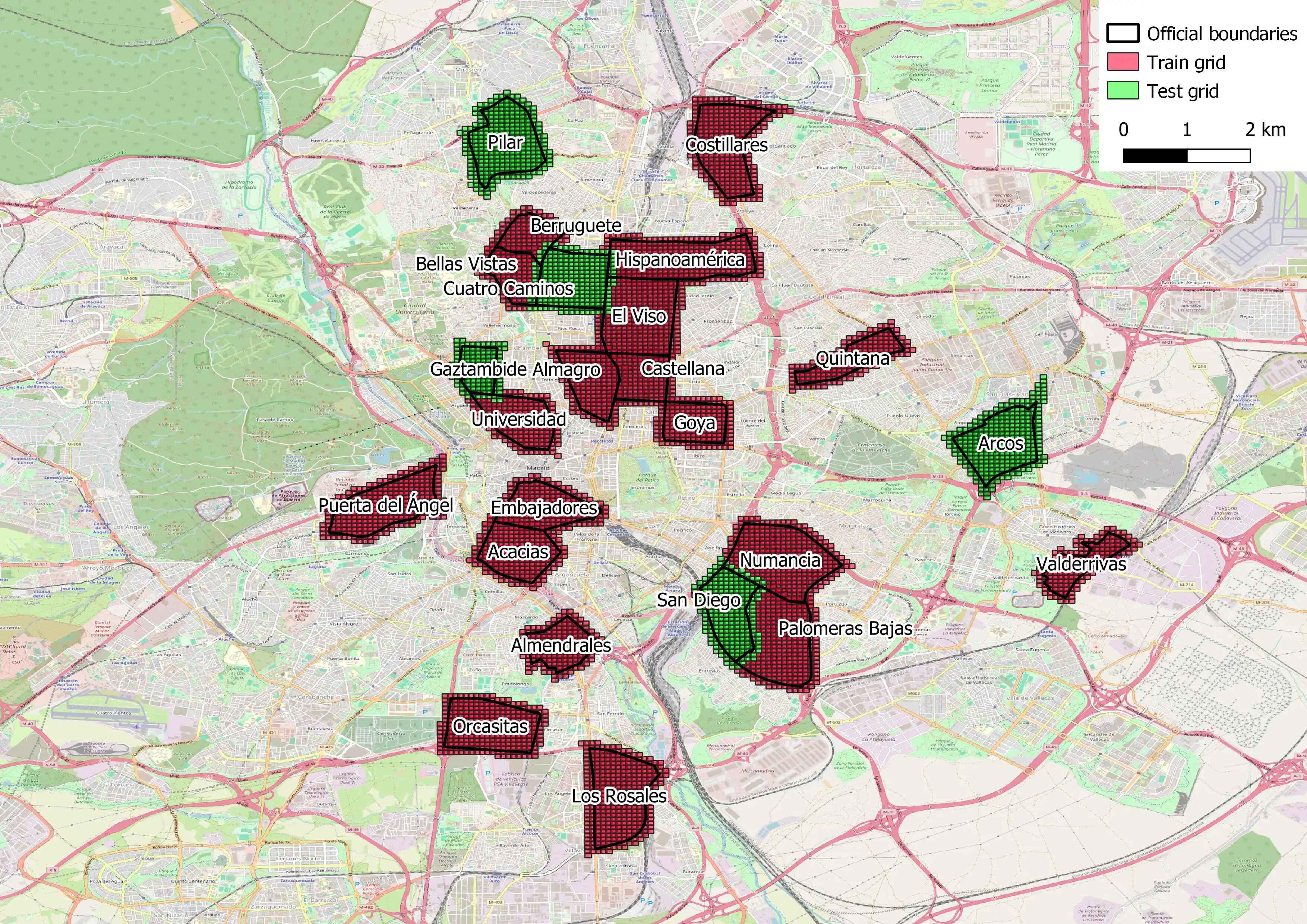}
            \caption{Training and testing neighbourhoods.}
            \label{fig:MadGrid}
        \end{subfigure} &
        \begin{subfigure}[t]{0.47\linewidth}
            \centering
            \includegraphics[width=5cm, height=4cm]{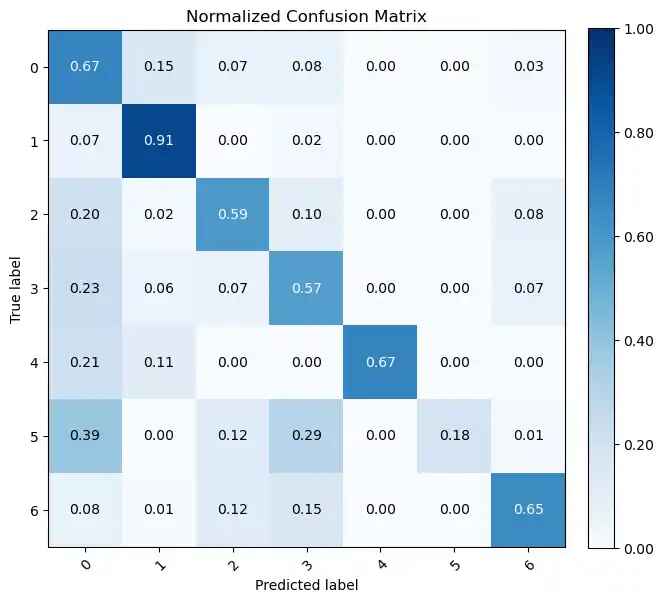}
            \caption{Normalized confusion matrix for the Madrid test set (Class 0 is background).}
            \label{fig:CMMadrid}
        \end{subfigure}
    \end{tabular}
    \caption{Madrid dataset and model results.}
    \label{fig:resMad}
\end{figure}

\begin{table}[!htb]
\caption{Evaluation results for the Madrid test set.}
\label{tab:madridtest}
\tiny
\begin{tabular}{|c|c|c|c|c|c|c|c|c|c|}
\hline
\textbf{\begin{tabular}[c]{@{}c@{}}Class \\ id\end{tabular}} & \textbf{Class name} & \textbf{\begin{tabular}[c]{@{}c@{}}model \\ area\end{tabular}} & \textbf{GT area} & \textbf{IoU} & \textbf{IoU\_200} & \textbf{F1} & \textbf{\begin{tabular}[c]{@{}c@{}}street \\ ratio\end{tabular}} & \textbf{\begin{tabular}[c]{@{}c@{}}pedestrian \\ ratio\end{tabular}} & \textbf{$\frac{model area}{GT   area}$} \\ \hline
1                                                            & building            & 2,769,163                                                    & 2,601,307      & 0.73         & 0.82              & 0.86        & 0.01                                                             & 0.02                                                                 & 1.07                                    \\ \hline
2                                                            & road                & 1,343,829                                                    & 1,776,690      & 0.50         & 0.55              & 0.67        & 0.81                                                             & 0.06                                                                 & 0.77                                    \\ \hline
3                                                            & sidewalk            & 1,103,500                                                    & 1,127,623      & 0.40         & 0.52              & 0.56        & 0.22                                                             & 0.57                                                                 & 0.99                                    \\ \hline
4                                                            & pool                & 21,403                                                       & 12,624         & 0.25         & 0.33              & 0.47        & 0.07                                                             & 0.02                                                                 & 2.61                                    \\ \hline
5                                                            & bike path           & 11,250                                                       & 18,060         & 0.06         & 0.08              & 0.19        & 0.18                                                             & 0.32                                                                 & 0.75                                    \\ \hline
6                                                            & parking             & 541,439                                                      & 438,223        & 0.36         & 0.52              & 0.55        & 0.75                                                             & 0.14                                                                 & 1.26                                    \\ \hline
\end{tabular}
\end{table}

The building class shows the highest accuracy among all classes. For the rest of the classes, the IoU and F1 score is in a medium to high range [fig \ref{fig:resMad} and tab \ref{tab:madridtest}]. However IoU\_200 indicates that the accuracy improves significantly if a deviation of a few centimeters or meters around the ground truth is allowed. The street and pedestrian ratios reveal minor errors in distinguishing between infrastructure intended for vehicles and pedestrians. Most inaccuracies occur between classes within similar categories. 

The bike lane and swimming pool classes have the lowest accuracies [fig \ref{fig:resMad} and tab \ref{tab:madridtest}]. However, the area covered by these classes in the ground truth is minimal or none in many training and testing areas, leading to inconsistent results for these classes. Nonetheless, as explained in the following section, correct trends over time were observed for all classes, including pool and bike lane. Overall, the model's results demonstrate that it is effective for predicting exact building boundaries and providing general statistical data for the other classes.

Even though there can be significant differences (over 15\%) in the area picked up by the model in comparison with ground truth for some classes, the assumption is made that the mistakes or differences between ground truth and model will not occur between the results of the same model evaluated with different images. For the comparison between 2001 and 2023, the model will apply the same criteria, overrepresenting or underrepresenting the classes in the images from both times by the same amount, allowing for a valid comparison. This is the reason why the model's output from 2001 is not compared directly to the ground truth from 2023.

\begin{figure}[!htb]
    \centering
    \begin{tabular}{ccccc}
        \begin{subfigure}[t]{0.19\linewidth}
            \centering
            \includegraphics[width=\linewidth]{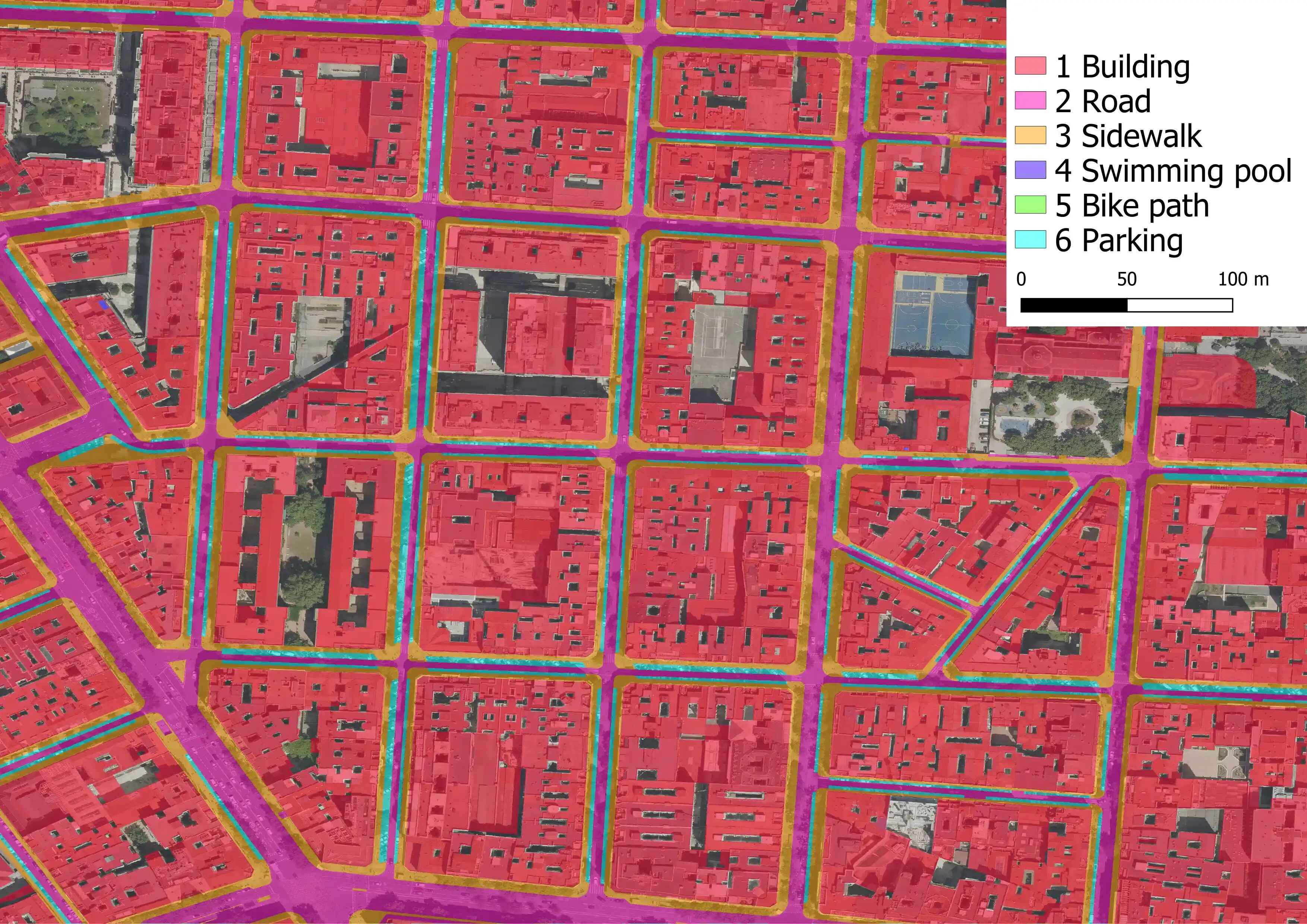}
            \caption{Gaztambide ground truth}
            \label{fig:gt_Gaztambide}
        \end{subfigure} &
        \begin{subfigure}[t]{0.19\linewidth}
            \centering
            \includegraphics[width=\linewidth]{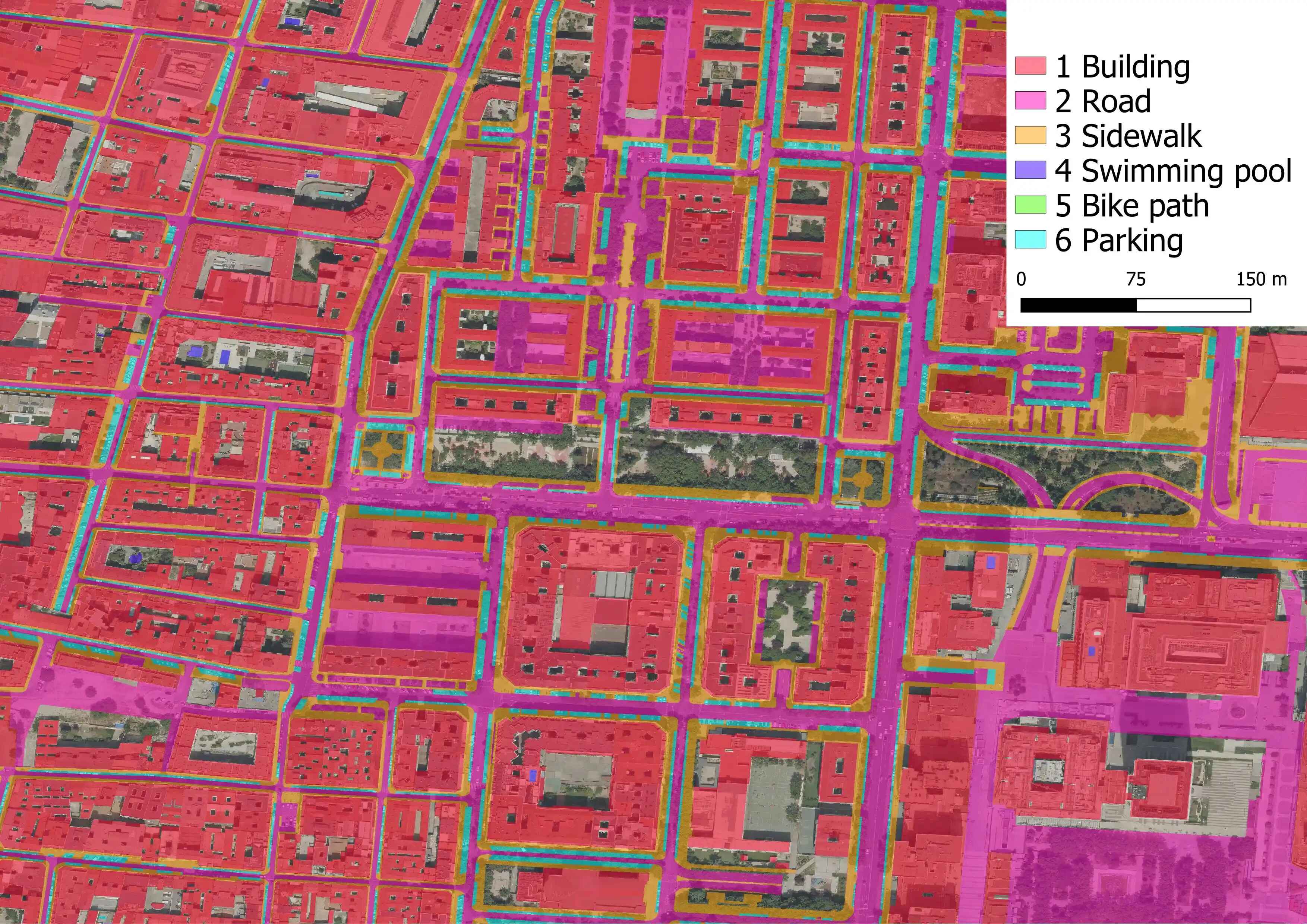}
            \caption{Cuatro Caminos ground truth}
            \label{fig:gt_cc}
        \end{subfigure} &
        \begin{subfigure}[t]{0.19\linewidth}
            \centering
            \includegraphics[width=\linewidth]{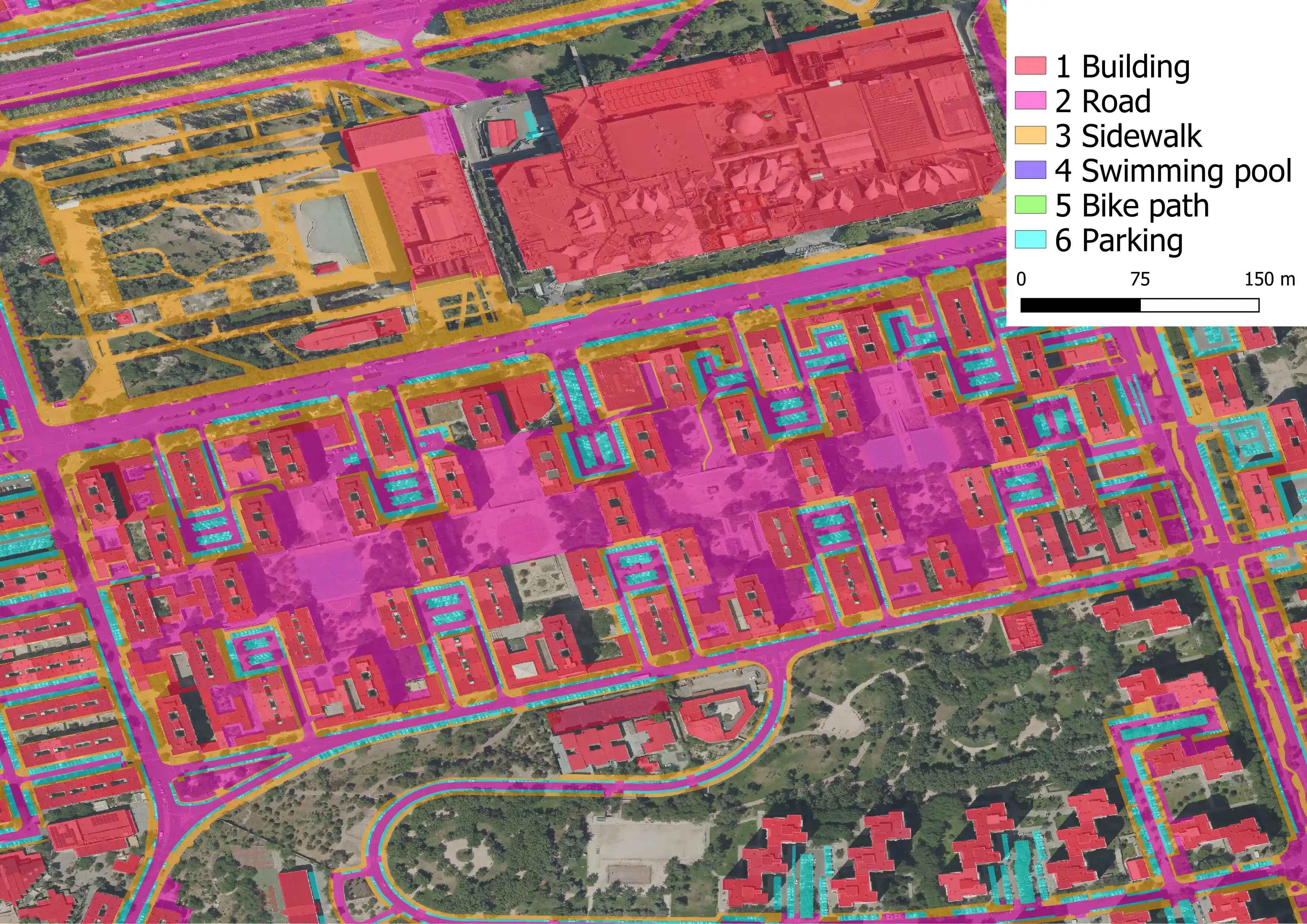}
            \caption{Pilar ground truth}
            \label{fig:gt_Pilar}
        \end{subfigure} &
        \begin{subfigure}[t]{0.19\linewidth}
            \centering
            \includegraphics[width=\linewidth]{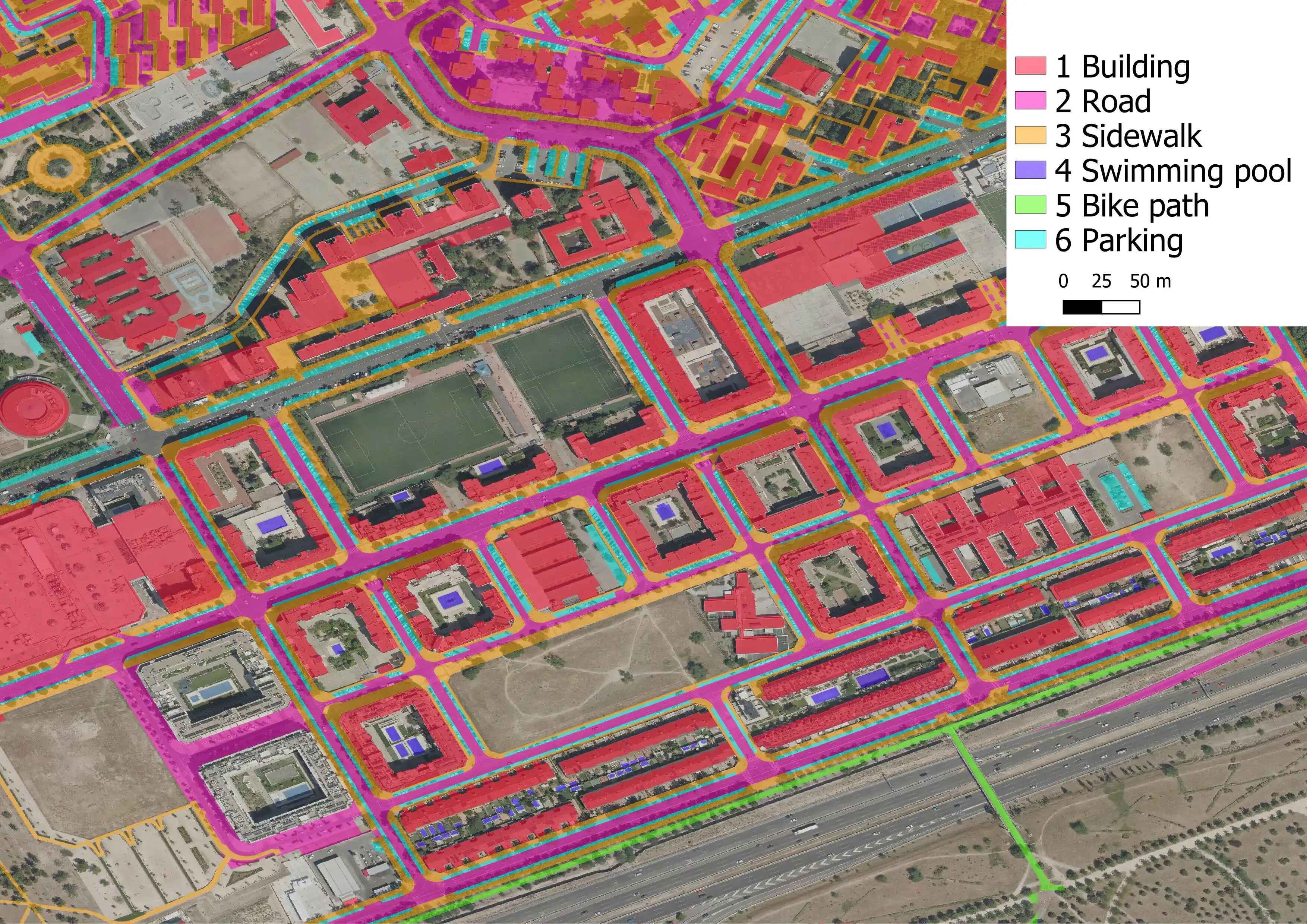}
            \caption{Arcos ground truth}
            \label{fig:gt_Arcos}
        \end{subfigure} &
        \begin{subfigure}[t]{0.19\linewidth}
            \centering
            \includegraphics[width=\linewidth]{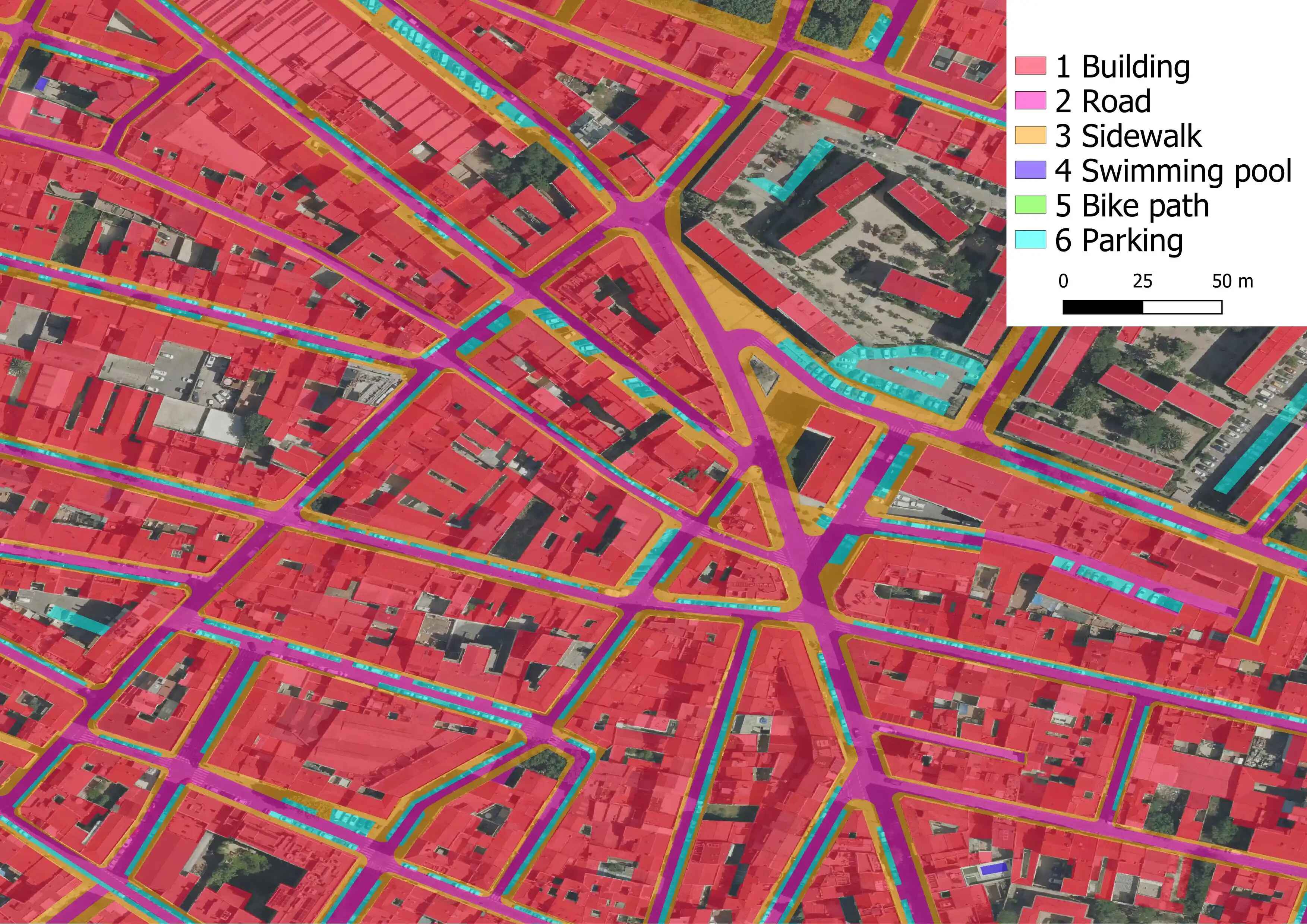}
            \caption{San Diego ground truth}
            \label{fig:gt_sdiego}
        \end{subfigure} \\
        
        \begin{subfigure}[t]{0.19\linewidth}
            \centering
            \includegraphics[width=\linewidth]{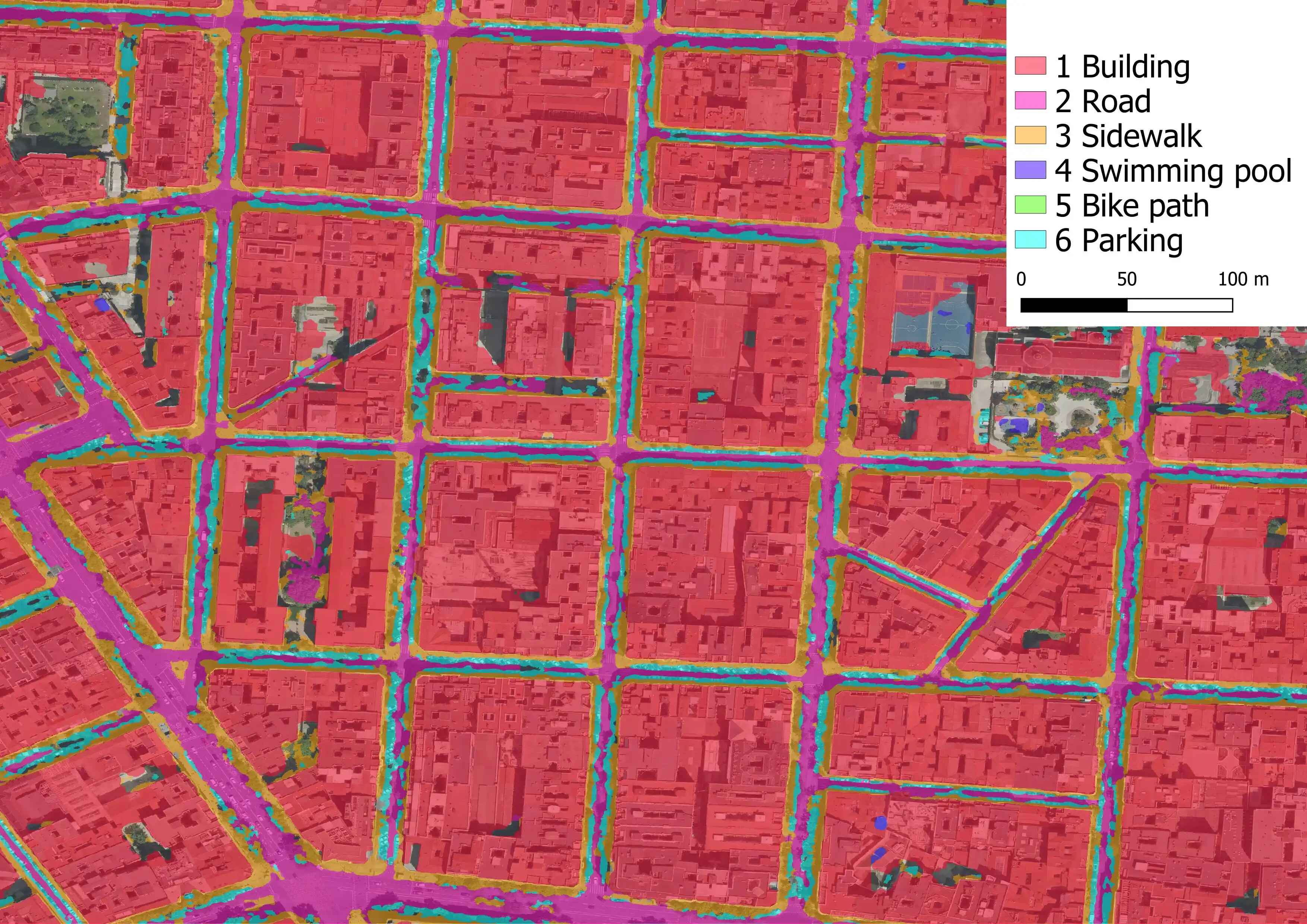}
            \caption{Gaztambide model 2023}
            \label{fig:2023_Gaztambide}
        \end{subfigure} &
        \begin{subfigure}[t]{0.19\linewidth}
            \centering
            \includegraphics[width=\linewidth]{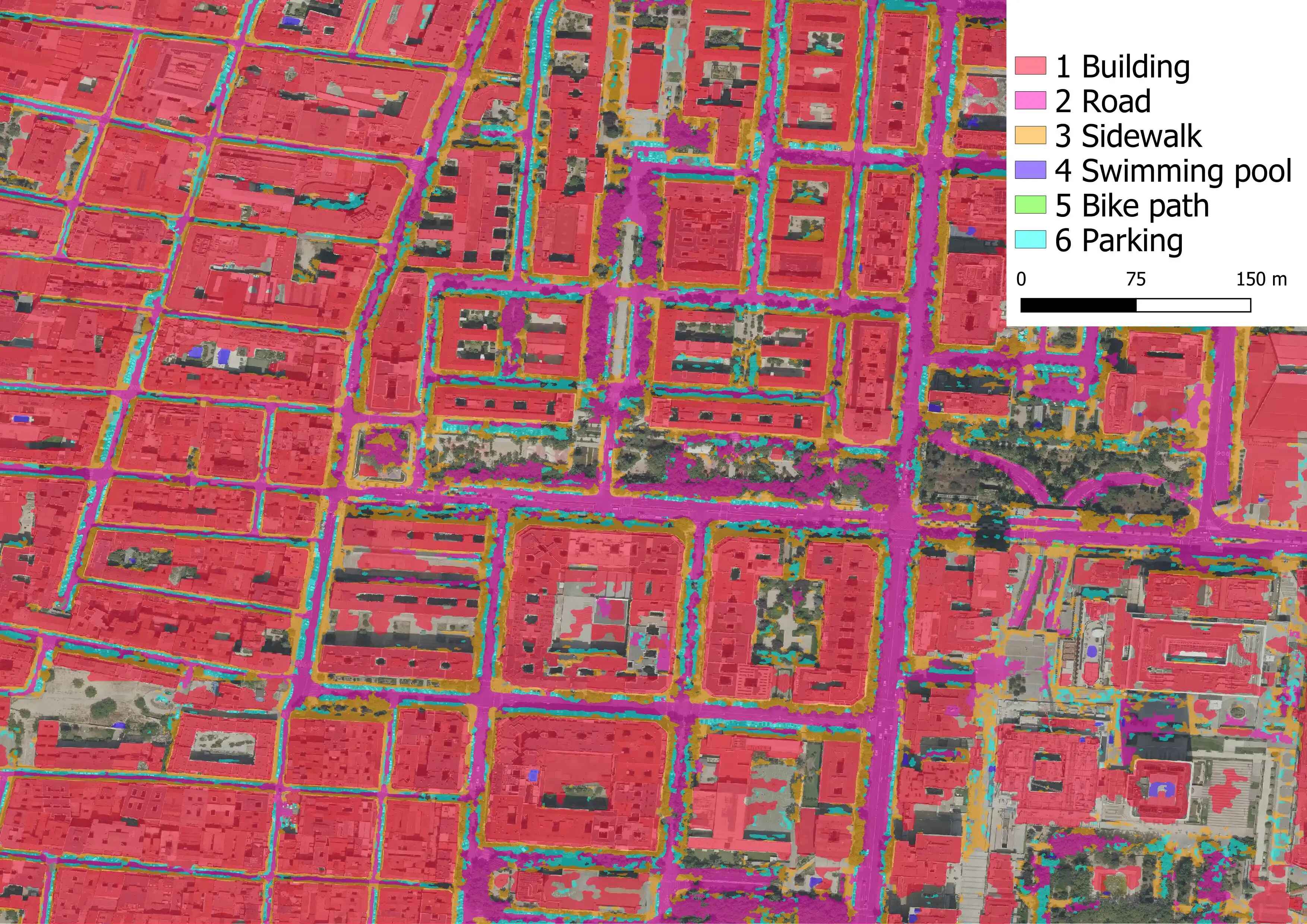}
            \caption{Cuatro Caminos model 2023}
            \label{fig:2023_cc}
        \end{subfigure} &
        \begin{subfigure}[t]{0.19\linewidth}
            \centering
            \includegraphics[width=\linewidth]{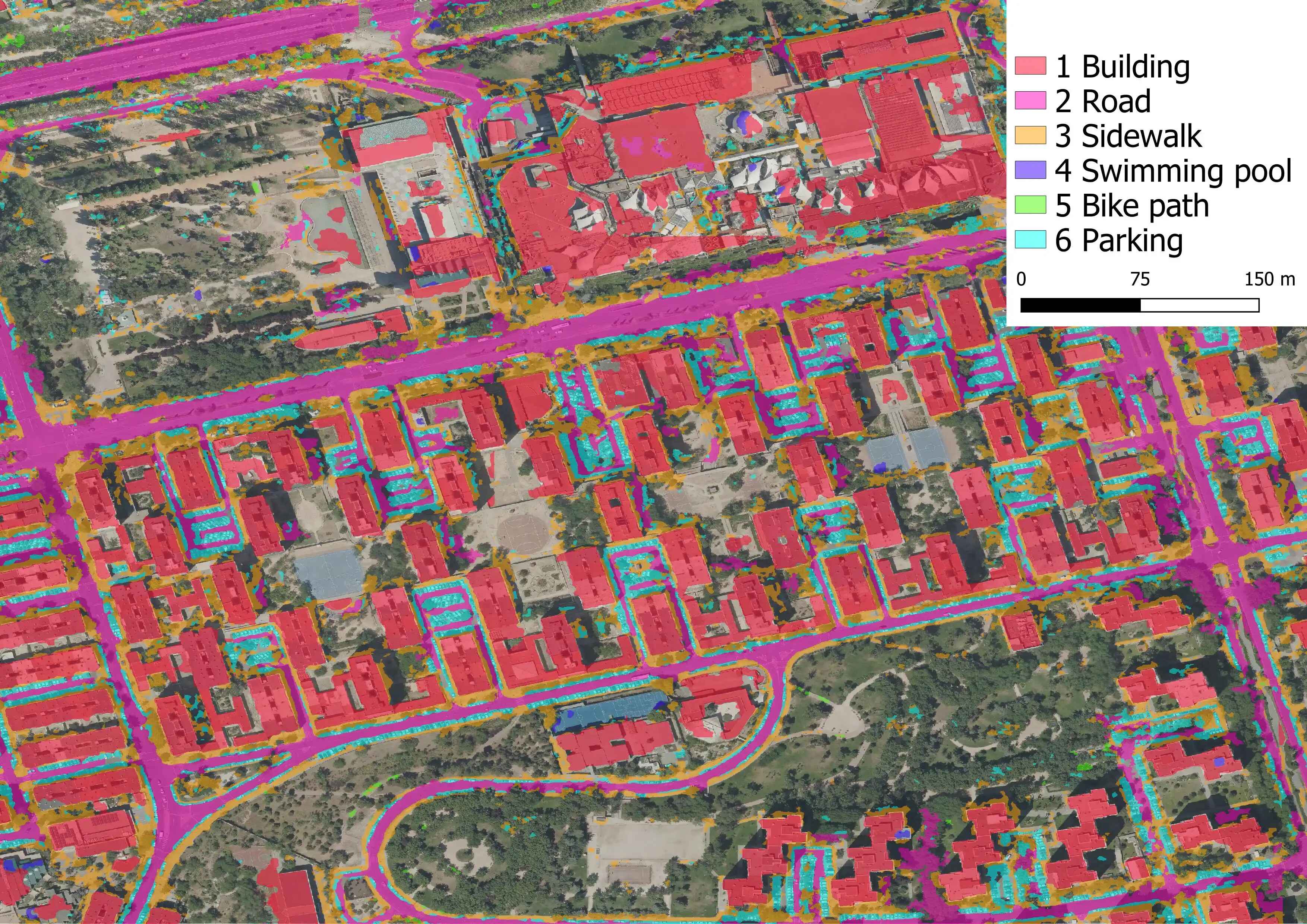}
            \caption{Pilar model 2023}
            \label{fig:2023_Pilar}
        \end{subfigure} &
        \begin{subfigure}[t]{0.19\linewidth}
            \centering
            \includegraphics[width=\linewidth]{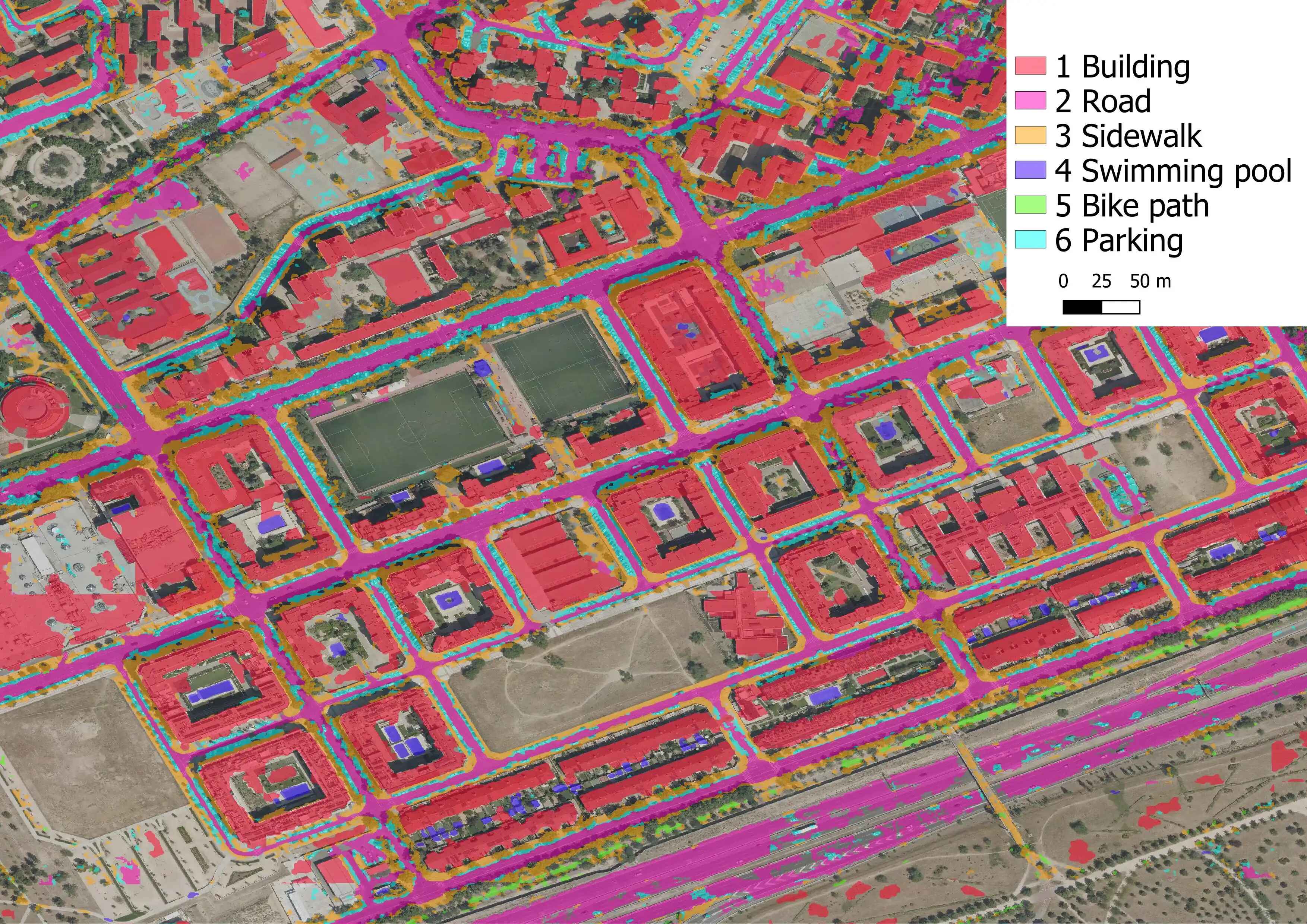}
            \caption{Arcos model 2023}
            \label{fig:2023_Arcos}
        \end{subfigure} &
        \begin{subfigure}[t]{0.19\linewidth}
            \centering
            \includegraphics[width=\linewidth]{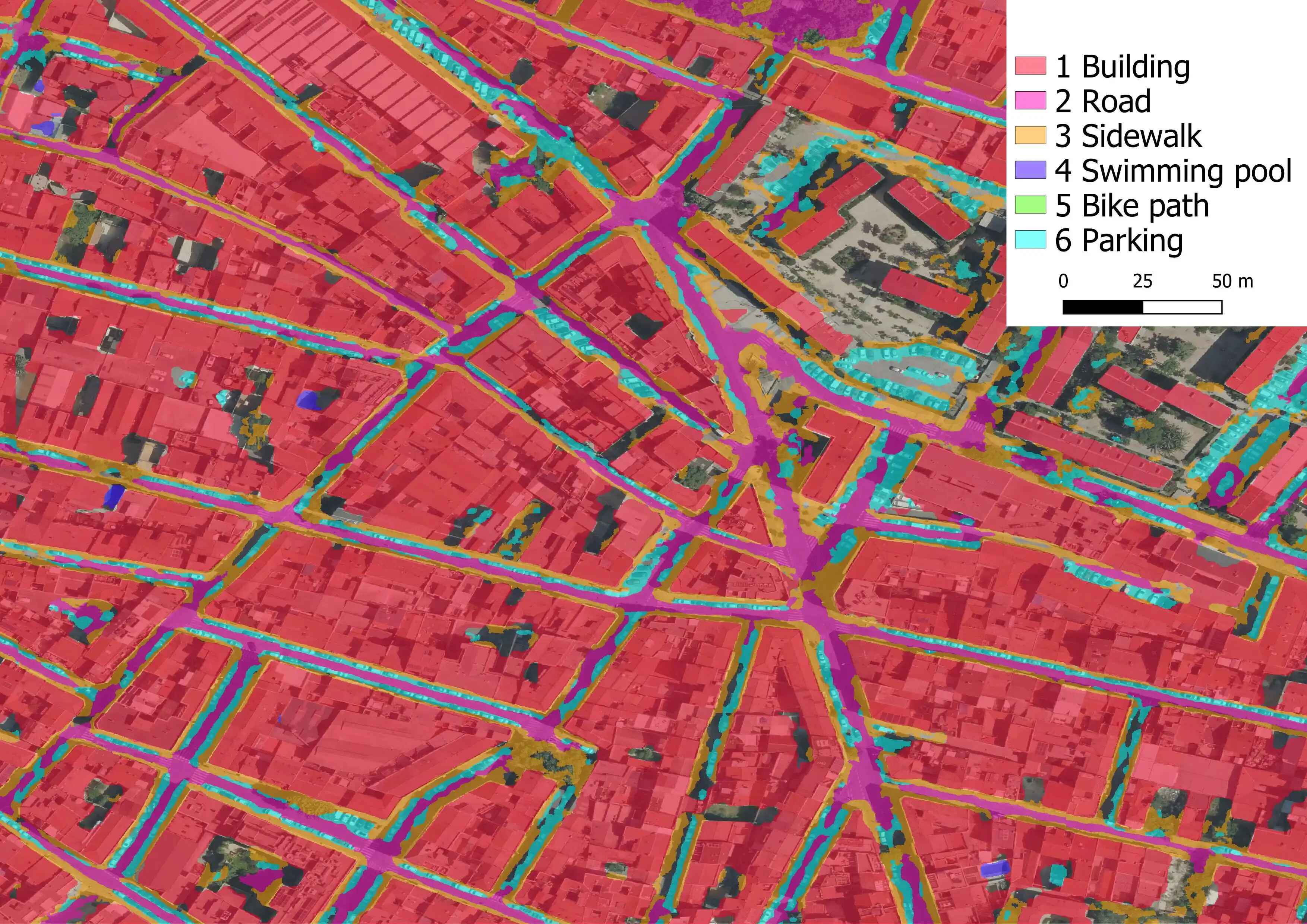}
            \caption{San Diego model 2023}
            \label{fig:2023_sdiego}
        \end{subfigure} \\
        
        \begin{subfigure}[t]{0.19\linewidth}
            \centering
            \includegraphics[width=\linewidth]{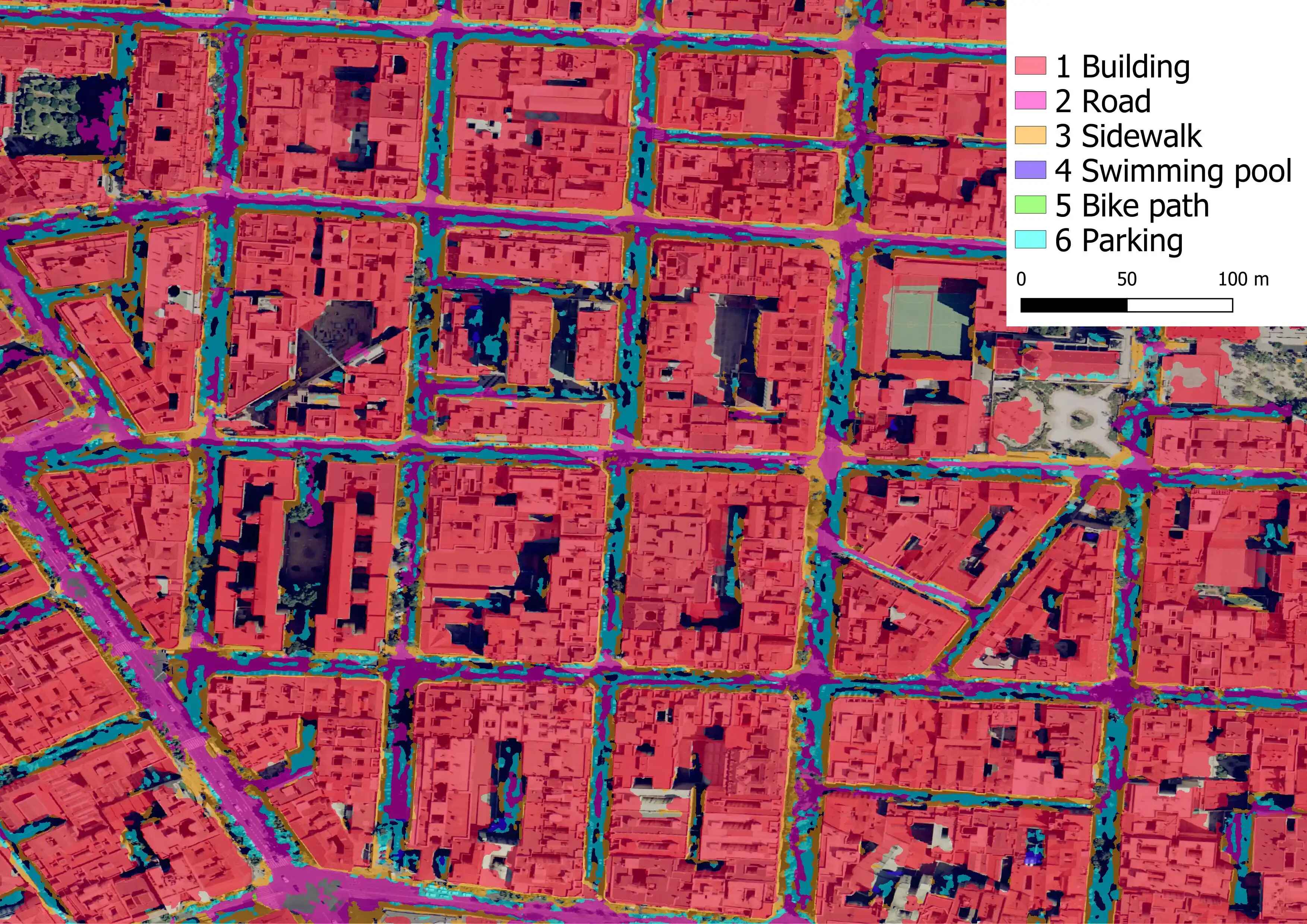}
            \caption{Gaztambide model 2001}
            \label{fig:2001_Gaztambide}
        \end{subfigure} &
        \begin{subfigure}[t]{0.19\linewidth}
            \centering
            \includegraphics[width=\linewidth]{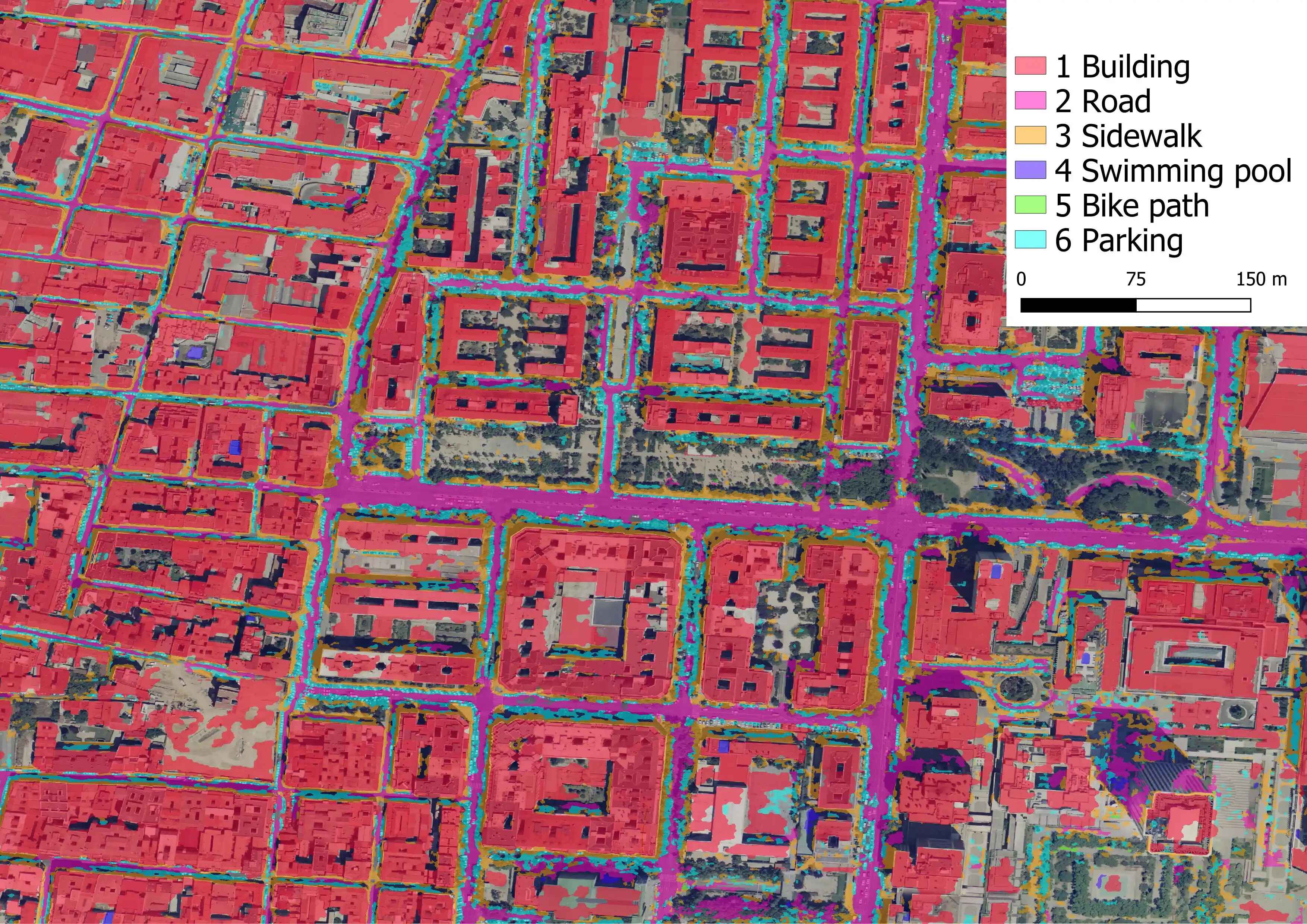}
            \caption{Cuatro Caminos model 2001}
            \label{fig:2001_cc}
        \end{subfigure} &
        \begin{subfigure}[t]{0.19\linewidth}
            \centering
            \includegraphics[width=\linewidth]{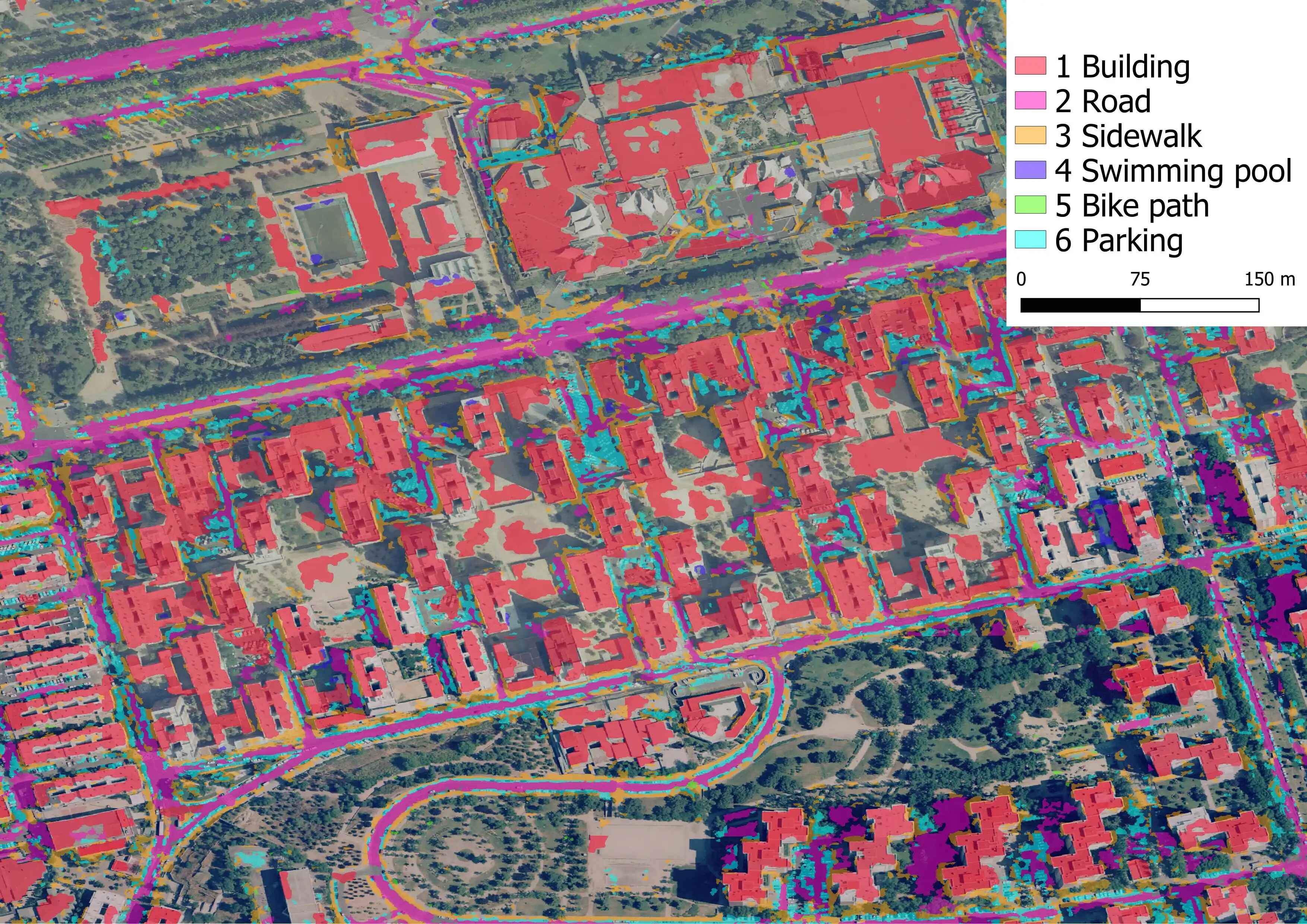}
            \caption{Pilar model 2001}
            \label{fig:2001_Pilar}
        \end{subfigure} &
        \begin{subfigure}[t]{0.19\linewidth}
            \centering
            \includegraphics[width=\linewidth]{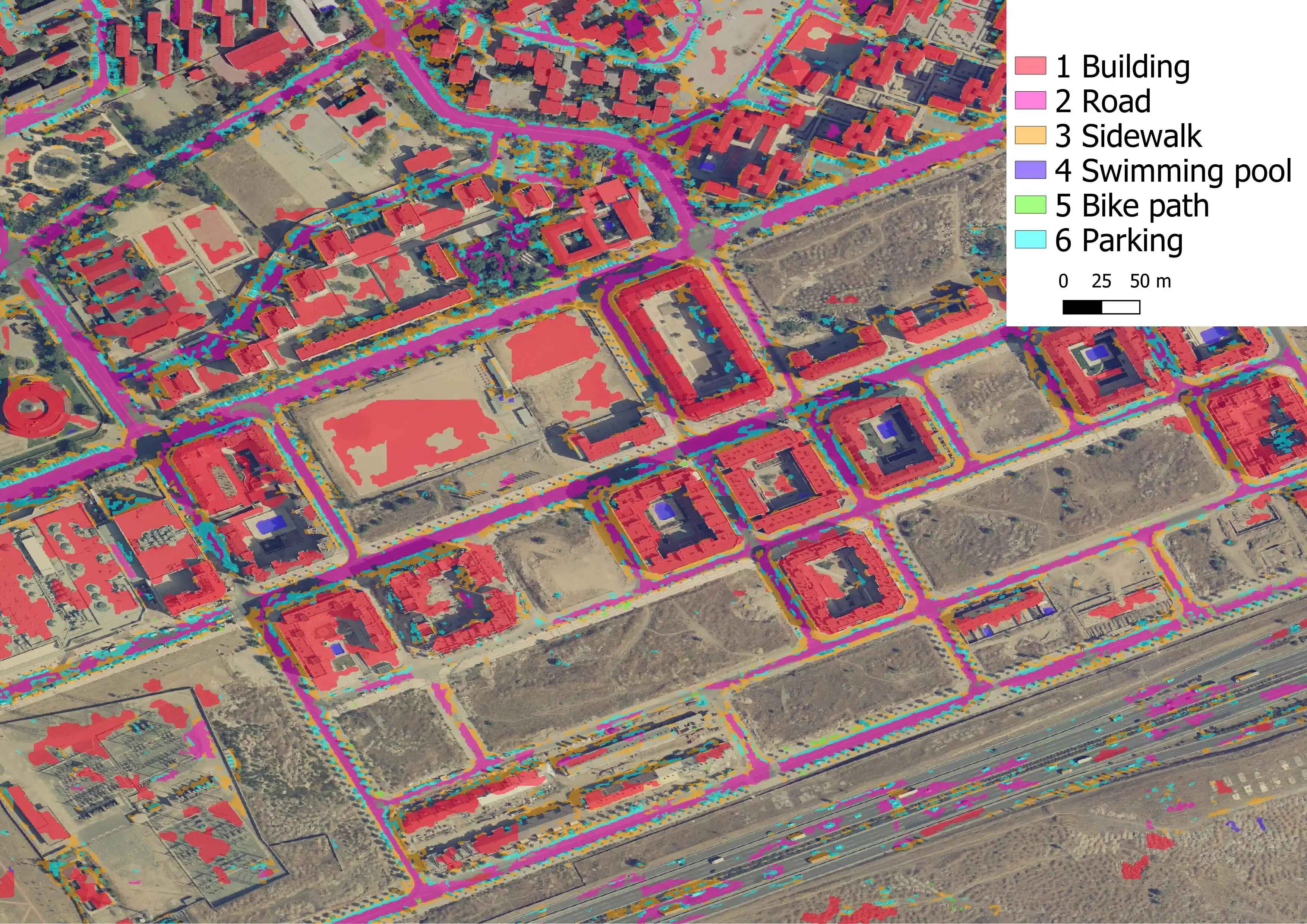}
            \caption{Arcos model 2001}
            \label{fig:2001_Arcos}
        \end{subfigure} &
        \begin{subfigure}[t]{0.19\linewidth}
            \centering
            \includegraphics[width=\linewidth]{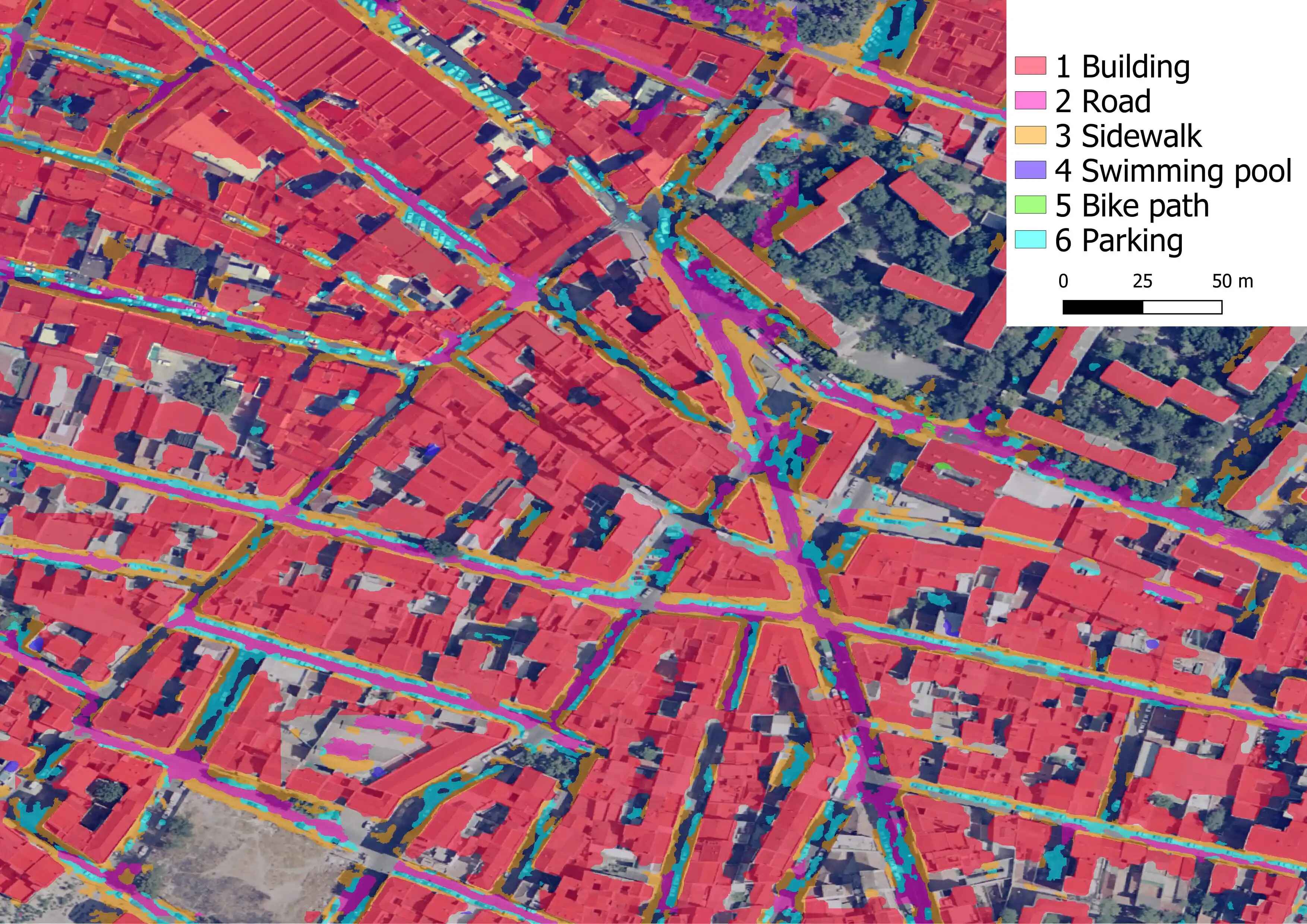}
            \caption{San Diego model 2001}
            \label{fig:2001_sdiego}
        \end{subfigure}
    \end{tabular}
    \caption{Madrid test dataset. An example region is shown for each testing neighbourhood. Ground truth from 2023 and model output for images from 2023 and 2001 are shown.}
    \label{fig:madridtest}
\end{figure}

\subsubsection{Temporal trends (2001-2023)}

Images from the same testing neighborhoods but from the year 2001 (RGB images with 10 cm per pixel resolution) were inputted into the model that was trained with images from 2023. The model's output from 2001 was compared to the model's output from 2023, evaluated previously. The IoU values indicate if the classes overlap and show the amount of variation that occurred during the timeframe. Very low IoU values (below 0.1), especially if the areas have not changed much, can indicate that the model is not providing accurate predictions for that class. The change in area shows the trend over time.

\begin{table}[!htb]
\centering
\tiny
\caption{Temporal trend (area in 2023 / area in 2001) for the Madrid test dataset and iou\_200 (in red) between the 2001 and 2023 geometries to show the validity of the results.}
\label{tab:mad2001}
\begin{tabular}{|c|c|c|c|c|c|c|}
    \hline
    \textbf{Dataset} & \textbf{1 building} & \textbf{2 road} & \textbf{3 sidewalk} & \textbf{4 pool} & \textbf{5 bike Path} & \textbf{6 parking} \\
    \hline
    Gaztambide & 1.11 \textcolor{red}{(0.80)} & 1.45 \textcolor{red}{(0.51)} & 1.27 \textcolor{red}{(0.46)} & 0.87 \textcolor{red}{(0.08)} & 0.11 \textcolor{red}{(0.00)} & 0.81 \textcolor{red}{(0.48)} \\
    \hline
    C. Caminos & 1.13 \textcolor{red}{(0.71)} & 1.36 \textcolor{red}{(0.57)} & 1.22 \textcolor{red}{(0.45)} & 1.07 \textcolor{red}{(0.10)} & 1.00 \textcolor{red}{(0.00)} & 0.79 \textcolor{red}{(0.38)} \\
    \hline
    Pilar & 1.15 \textcolor{red}{(0.52)} & 1.67 \textcolor{red}{(0.41)} & 1.65 \textcolor{red}{(0.29)} & 1.31 \textcolor{red}{(0.23)} & 2.17 \textcolor{red}{(0.02)} & 1.40 \textcolor{red}{(0.31)} \\
    \hline
    Arcos & 1.22 \textcolor{red}{(0.53)} & 1.64 \textcolor{red}{(0.46)} & 1.54 \textcolor{red}{(0.32)} & 2.04 \textcolor{red}{(0.20)} & 2.88 \textcolor{red}{(0.01)} & 1.41 \textcolor{red}{(0.31)} \\
    \hline
    San Diego & 1.10 \textcolor{red}{(0.69)} & 1.85 \textcolor{red}{(0.41)} & 1.41 \textcolor{red}{(0.41)} & 1.02 \textcolor{red}{(0.08)} & 2.01 \textcolor{red}{(0.05)} & 1.34 \textcolor{red}{(0.35)} \\
    \hline
    ALL & 1.13 \textcolor{red}{(0.65)} & 1.58 \textcolor{red}{(0.47)} & 1.42 \textcolor{red}{(0.39)} & 1.34 \textcolor{red}{(0.14)} & 2.35 \textcolor{red}{(0.02)} & 1.15 \textcolor{red}{(0.37)} \\
    \hline
\end{tabular}
\end{table}

In the center of the city [tab \ref{tab:mad2001}], specifically in the Gaztambide and Cuatro Caminos areas, the trend for parking spaces is decreasing, by around 20\%. However, there is an increase in the surface of roads and sidewalks, with a growth of 30 to 40\%. The number of buildings remains mostly constant, with a slight increase of 10\%.

In the periphery of the city [tab \ref{tab:mad2001}], including Pilar, Arcos, and San Diego, the trends for parking, roads, and sidewalks are increasing, with about a 50\% increase in surface area. Building growth varies, with a slight increase of 10 to 15\% in Pilar and San Diego, and a moderate increase of 22\% in Arcos. Arcos, the newest neighborhood developed in the early 2000s, shows a significant increase in the number of swimming pools (104\%), a trend not seen in other neighborhoods, where the increase is less than 30\%. This is due to the tendency at that time to build multifamily housing with shared swimming pools in new developments.

There are no significant differences in the trends in road and parking surface between richer neighborhoods, such as Pilar, and poorer neighborhoods, such as San Diego. The most notable differences are observed between the city's periphery and its center.

\subsection{Vienna}

Public road and tram tracks show the best accuracies. For the rest of the classes the IoU is lower, but the rest of the indices can validate the results [tab \ref{tab:viennatest}]. IoU\_200 (23-GT calculated with ground truth and model results for 2023 images) show that the accuracy increases vastly if a deviation of a few centimeters or meters around the ground truth is tolerated. The street and pedestrian ratios show small amounts of errors between detecting infrastructure for cars or for pedestrians. Most of the inaccuracies arise between similar looking classes, like parking and road. 

\begin{table}[!htb]
\tiny
\caption{Evaluation metrics and temporal trend (2014-2023) for the Vienna model.}
\label{tab:viennatest}
\begin{tabular}{|c|c|c|c|c|c|c|c|c|c|c|c|}
\hline
\textbf{\begin{tabular}[c]{@{}c@{}}Class \\ id\end{tabular}} & \textbf{\begin{tabular}[c]{@{}c@{}}Class \\ name\end{tabular}} & \textbf{\begin{tabular}[c]{@{}c@{}}model \\ area\end{tabular}} & \textbf{\begin{tabular}[c]{@{}c@{}}GT \\ area\end{tabular}} & \textbf{iou} & \textbf{\begin{tabular}[c]{@{}c@{}}iou\_200 \\ (23-GT)\end{tabular}} & \textbf{F1} & \textbf{\begin{tabular}[c]{@{}c@{}}street \\ ratio\end{tabular}} & \textbf{\begin{tabular}[c]{@{}c@{}}pedes-\\ trian \\ ratio\end{tabular}} & \textbf{$\frac{model area}{GT   area}$} & \textbf{\begin{tabular}[c]{@{}c@{}}iou\_200   \\ (14-23)\end{tabular}} & \textbf{$\frac{area 23}{area   14}$} \\ \hline
1                                                            & \begin{tabular}[c]{@{}c@{}}public \\ road\end{tabular}         & 163,676                                                        & 159,214                                                     & 0.47         & 0.56                                                                 & 0.69        & 0.79                                                             & 0.11                                                                     & 1.03                                    & 0.59                                                                   & 1.09                                 \\ \hline
2                                                            & \begin{tabular}[c]{@{}c@{}}rail \\ tracks\end{tabular}         & 31,696                                                         & 28,819                                                      & 0.55         & 0.66                                                                 & 0.76        & 0.09                                                             & 0.04                                                                     & 1.10                                    & 0.46                                                                   & 0.69                                 \\ \hline
3                                                            & crosswalk                                                      & 4,781                                                          & 2,999                                                       & 0.33         & 0.57                                                                 & 0.51        & 0.37                                                             & 0.52                                                                     & 1.59                                    & 0.40                                                                   & 1.37                                 \\ \hline
4                                                            & parking                                                        & 71,956                                                         & 50,593                                                      & 0.23         & 0.37                                                                 & 0.45        & 0.61                                                             & 0.13                                                                     & 1.42                                    & 0.35                                                                   & 1.61                                 \\ \hline
5                                                            & \begin{tabular}[c]{@{}c@{}}private \\ road\end{tabular}        & 157,414                                                        & 166,984                                                     & 0.32         & 0.36                                                                 & 0.51        & 0.55                                                             & 0.06                                                                     & 0.94                                    & 0.37                                                                   & 0.89                                 \\ \hline
6                                                            & sidewalk                                                       & 153,034                                                        & 97,993                                                      & 0.25         & 0.44                                                                 & 0.45        & 0.20                                                             & 0.40                                                                     & 1.56                                    & 0.42                                                                   & 1.33                                 \\ \hline
7                                                            & \begin{tabular}[c]{@{}c@{}}pedestrian \\ path\end{tabular}     & 195,103                                                        & 145,929                                                     & 0.20         & 0.33                                                                 & 0.37        & 0.08                                                             & 0.34                                                                     & 1.34                                    & 0.39                                                                   & 0.99                                 \\ \hline
\end{tabular}
\end{table}

\begin{figure}[!htb]
    \centering
    \begin{tabular}{ccc}
        \begin{subfigure}[t]{0.3\linewidth}
            \centering
            \includegraphics[width=\linewidth]{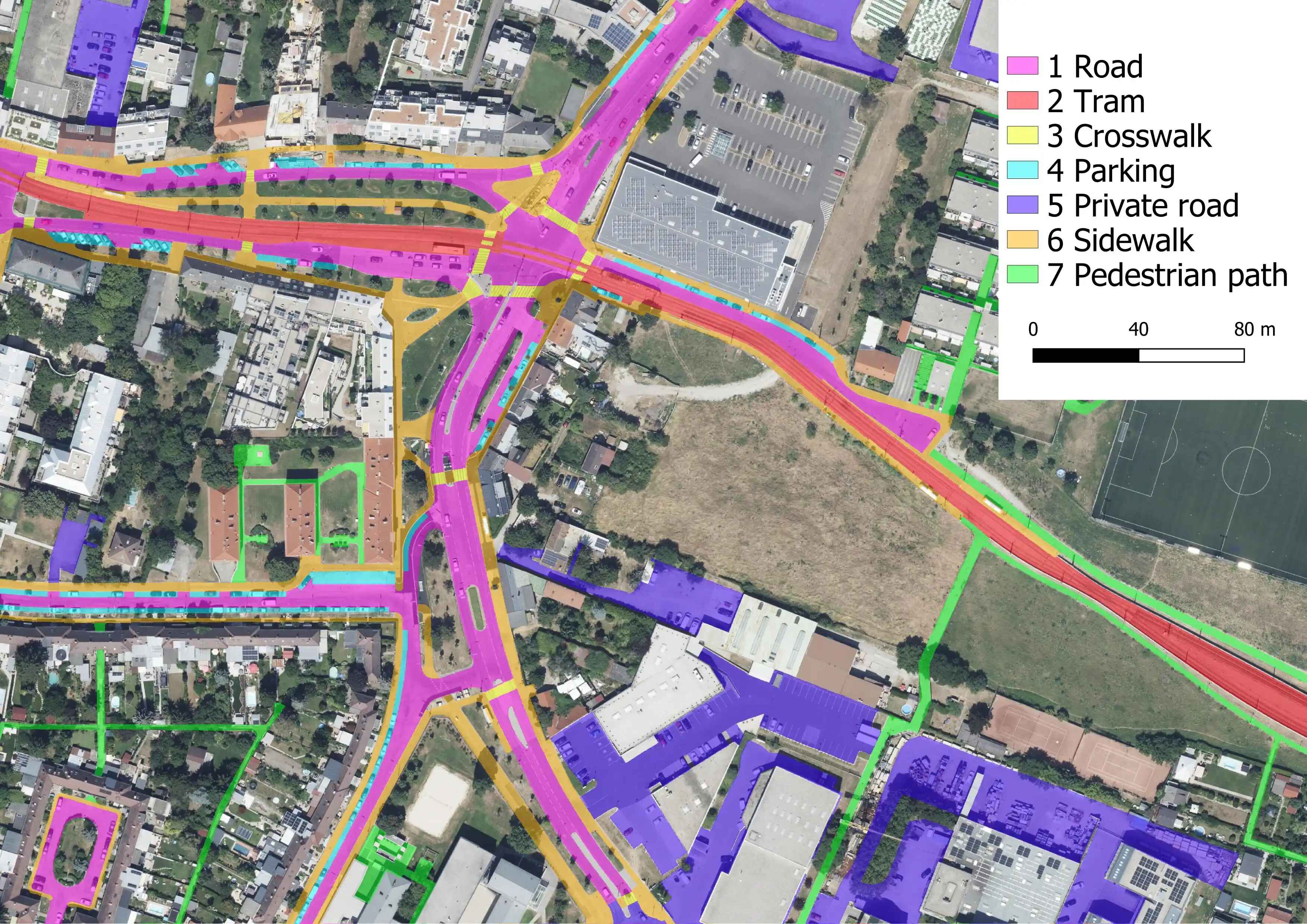}
            \caption{Vienna test set ground truth}
            \label{fig:gt_vienna_1}
        \end{subfigure} &
        \begin{subfigure}[t]{0.3\linewidth}
            \centering
            \includegraphics[width=\linewidth]{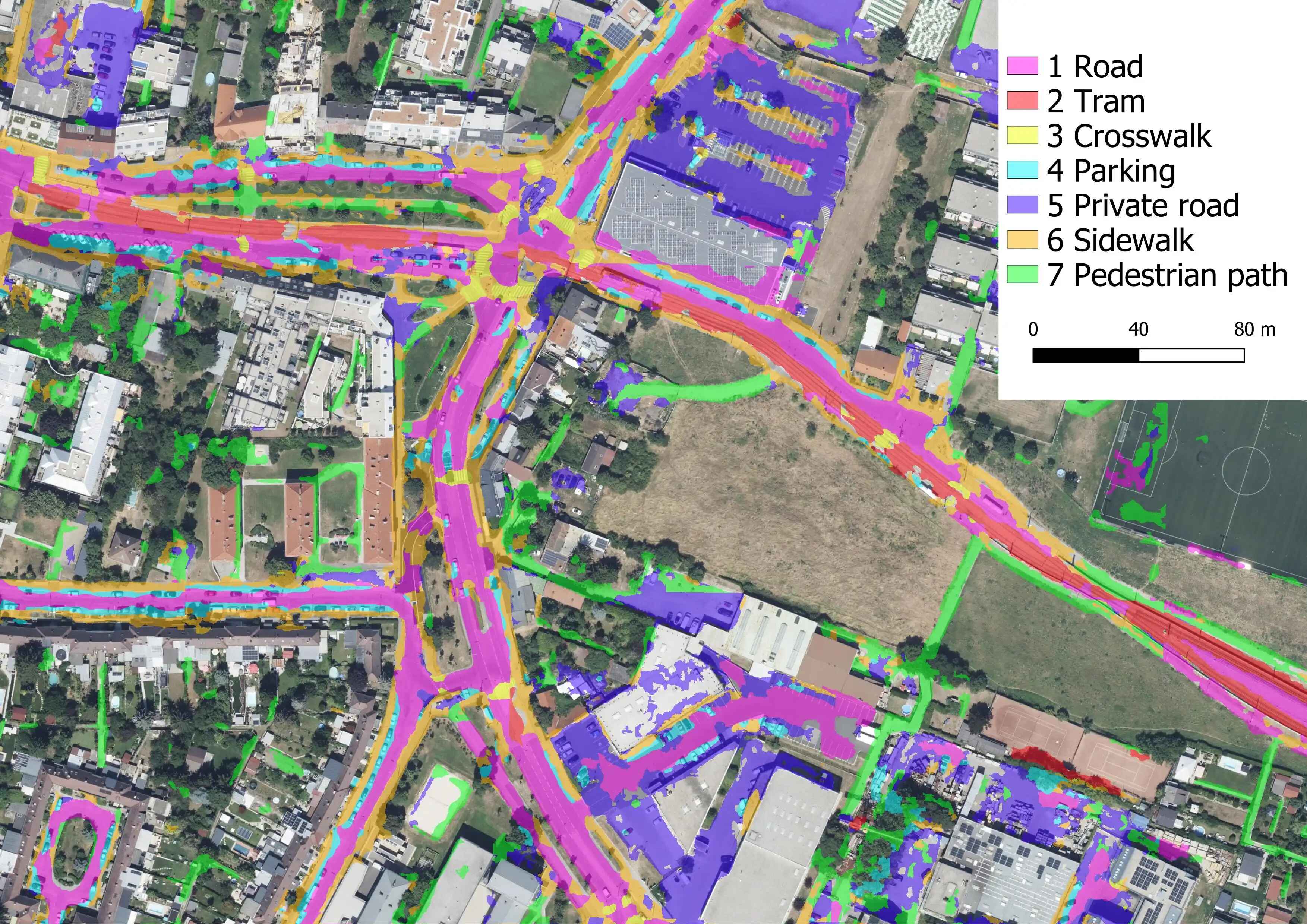}
            \caption{Model output for the 2023 image}
            \label{fig:2023_vienna_1}
        \end{subfigure} &
        \begin{subfigure}[t]{0.3\linewidth}
            \centering
            \includegraphics[width=\linewidth]{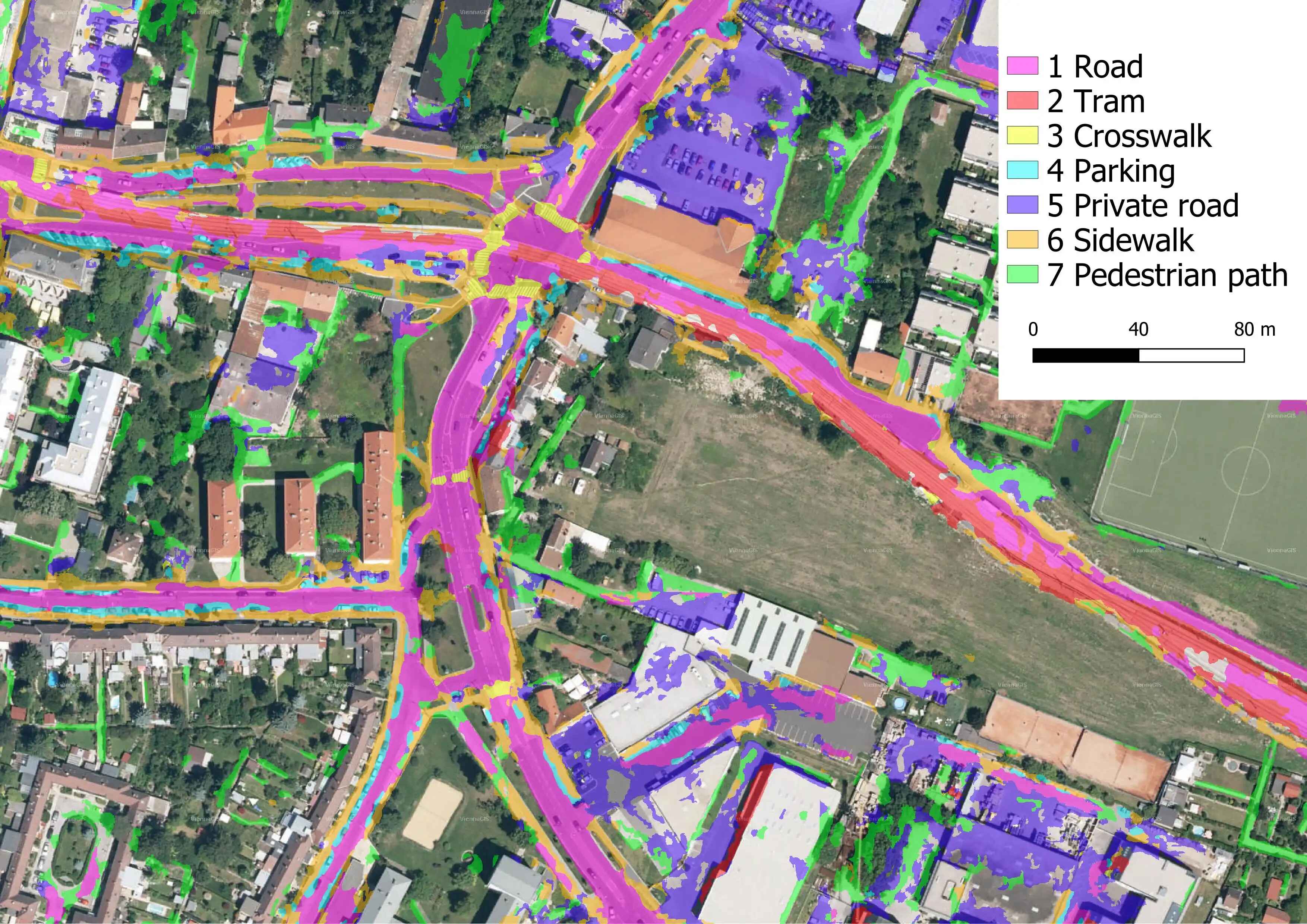}
            \caption{Model output for the 2014 image}
            \label{fig:2014_vienna_1}
        \end{subfigure} \\
        
        \begin{subfigure}[t]{0.3\linewidth}
            \centering
            \includegraphics[width=\linewidth]{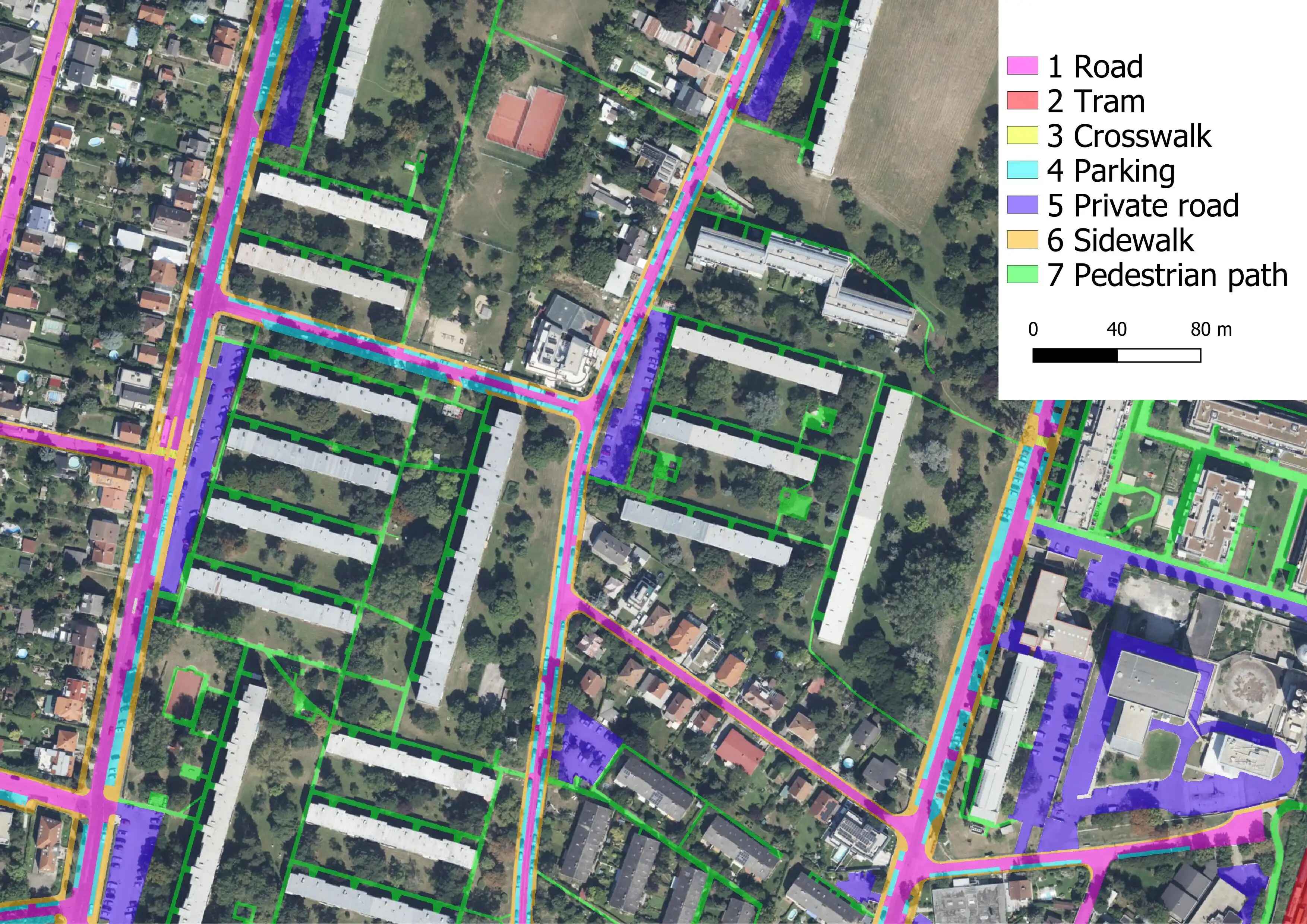}
            \caption{Vienna test set ground truth}
            \label{fig:gt_vienna_2}
        \end{subfigure} &
        \begin{subfigure}[t]{0.3\linewidth}
            \centering
            \includegraphics[width=\linewidth]{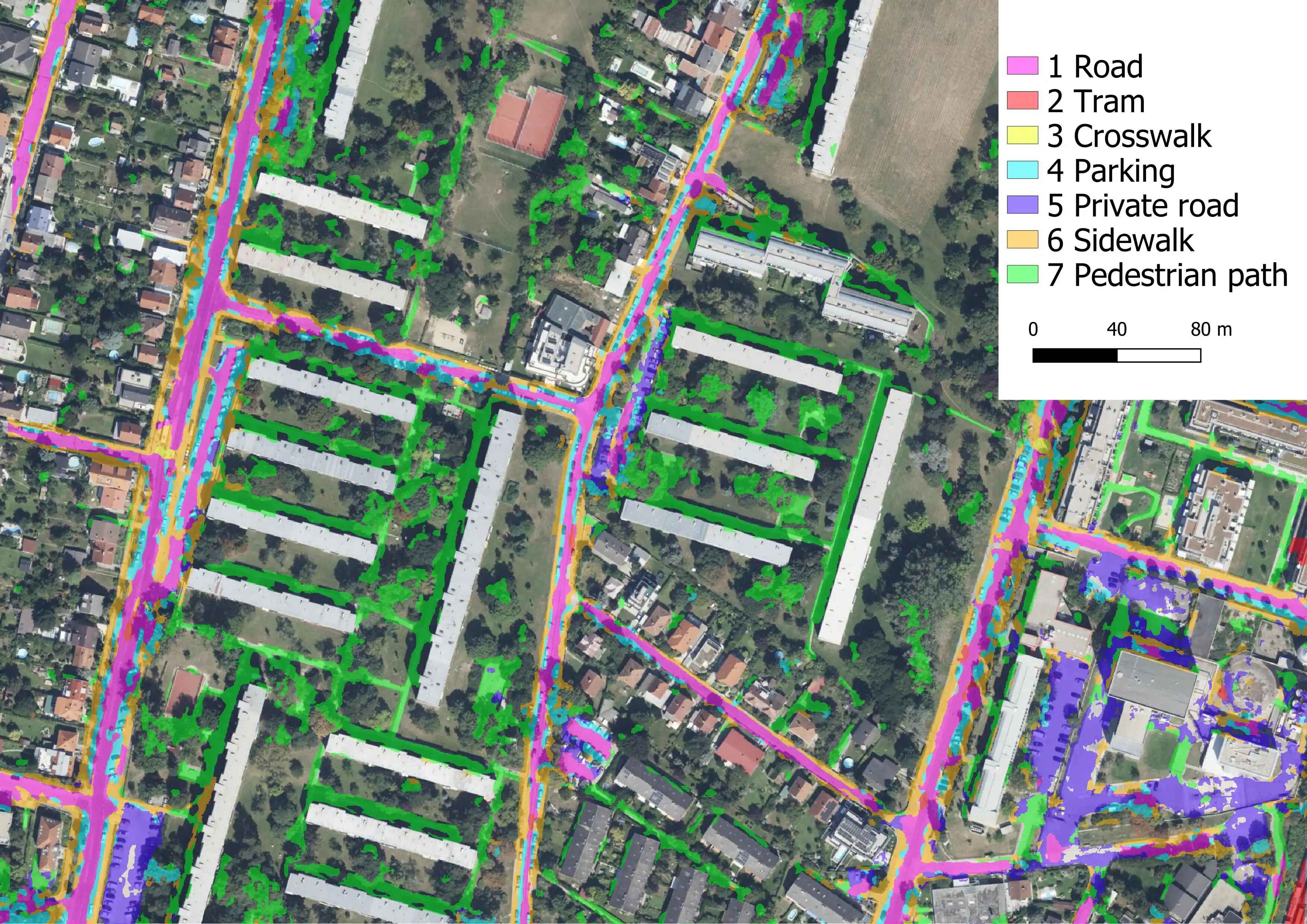}
            \caption{Model output for the 2023 image}
            \label{fig:2023_vienna_2}
        \end{subfigure} &
        \begin{subfigure}[t]{0.3\linewidth}
            \centering
            \includegraphics[width=\linewidth]{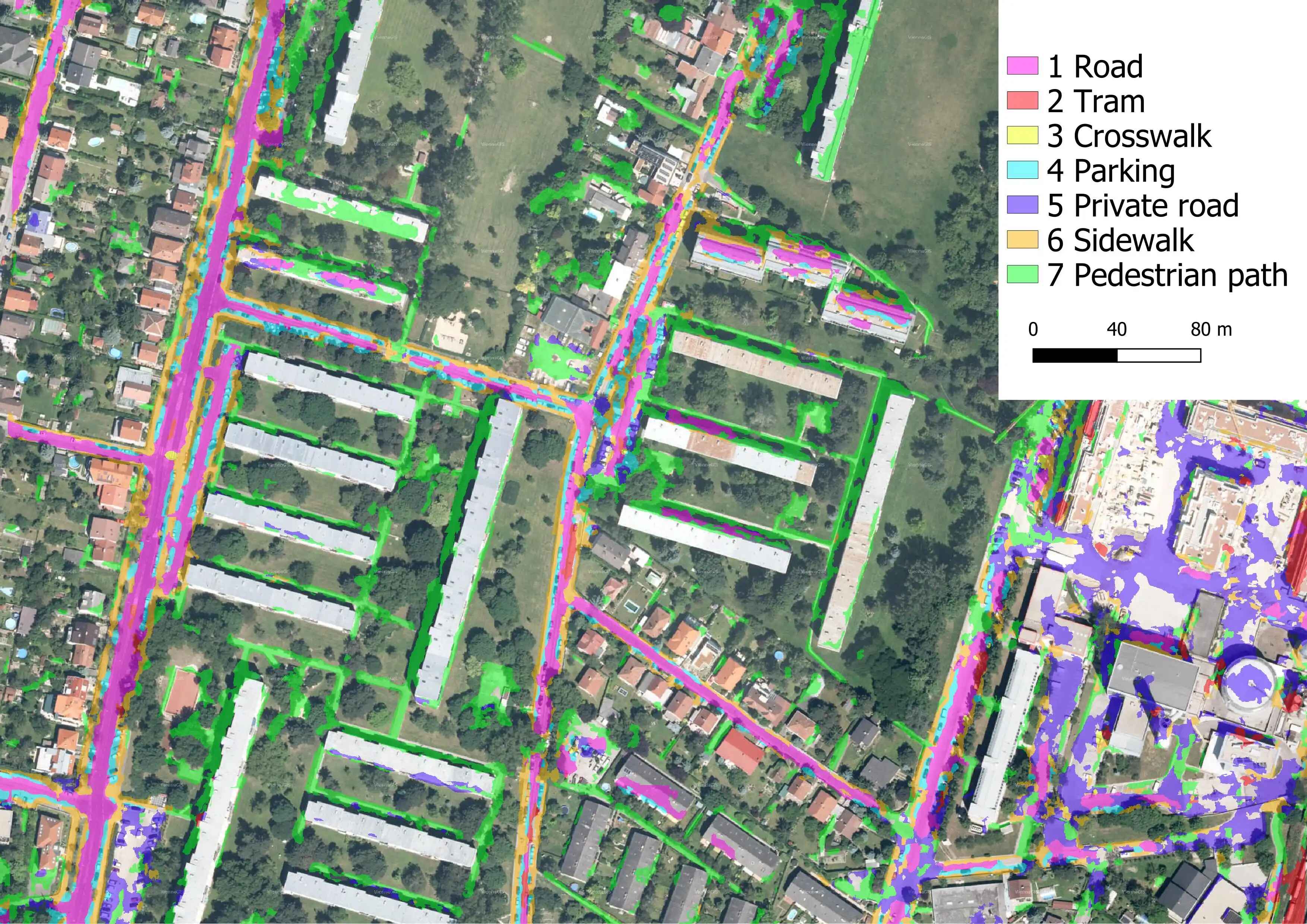}
            \caption{Model output for the 2014 image}
            \label{fig:2014_vienna_2}
        \end{subfigure} \\
    \end{tabular}
    \caption{Vienna test dataset. The two regions cover most of the test set. Ground truth from 2023, model output for images from 2023 and 2014.}
    \label{fig:viennatest}
\end{figure}

The model trained with 2023 imagery was tested with images from 2014, the first available RGB images. The outputs for both time periods are compared in table \ref{tab:viennatest} (columns iou\_200 (14-23 calculated with model results from 2014 and 2023) and $\frac{area 23}{area 14}$) and figure \ref{fig:viennatest}. Results show a decrease in tram tracks and private roads. The amount of road surfaces, bike paths, and pedestrian paths remains constant. Sidewalks increase, but the most significant growth is observed in public on-street parking, confirming the trend towards car-centric developments in the newer areas of Vienna.

To validate the transfer learning approach, a model was trained on this dataset using the SAM encoder, while another UNet model was trained without utilizing SAM's encoder weights. The transfer learning SAM model outperformed the UNet model in the mean IoU metric by 7\%, even though it was trained with only 10 epochs compared to the UNet's 30 epochs.

\subsection{Parking models}

Two models were trained using only the parking datasets, each with two classes. One model was trained with Madrid data and evaluated in Vienna, while the other model was trained in Vienna and evaluated in Madrid. The Madrid-trained model achieved an IoU of 0.18 and an IoU\_200 of 0.31 in the Vienna test set, which is a decrease of only 0.06 compared to the Vienna-trained model. Similarly, the Vienna-trained model achieved an IoU of 0.2 and an IoU\_200 of 0.3 in the Madrid test set. In that case the decrease was higher, 0.2, but it has to be considered that the IoU\_200 value for that model in Vienna was 0.37. This demonstrates the generalizability of the models and the validity of the method when working with images from cities that look very different. Images from different times but from the same city as the training set are more similar, so the results should be more accurate.

\section{Discussion}

Results indicate that the models are accurate enough to gather statistical data, such as the overall area in a neighborhood used for various purposes like housing, cars, parking, or pedestrians. The models exhibit sufficient generalizability to work in environments different from the training dataset as it was shown running tests on the Madrid-trained model in Vienna and on the Vienna-trained model in Madrid. It was possible to compare results from 2023 and 2014 for Vienna and from 2001 for Madrid, dates before government official data is available. This allowed us to compare trends over time and distinguish increases, decreases, and stable tendencies, though exact values exhibit errors if more than one decimal of precision is desired. The difference in the area between the models and ground truth is significant for some classes, with values over 15\%. However, as the ground truth can have missing data, overestimations in area might actually be correct. Unfortunately, it was not possible to provide exact estimates for the amount of over- or under-detection for the classes with the worst performance. To achieve this, a small dataset with manual labels could be created, but this is a very time-consuming task and is not desirable for the objectives of this study.

Overall, evaluating the precision of the models remains a challenge, as ground truth data is inaccurate. For certain classes, the models' results are accurate, but the ground truth is wrong, as shown in figures \ref{fig:madwrong} and \ref{fig:viewrong}. This issue makes testing results appear worse than those of other research \cite{henry_citywide_2021} (which reported IoU values around 0.6), where the test and train datasets were refined with extensive manual work. Image resolution appears to be an influential factor, as the Madrid-trained models with images at 10 cm per pixel achieved much better results than the Vienna-trained models with images at 15 cm per pixel.

OSM data proved to be valuable in creating datasets covering topics not included in official data, such as private parking surfaces in Vienna. However, evaluating the model remains a challenge. Without a high-quality test set, it is impossible to determine the validity of the model. Therefore, if OSM data is to be used, manual labeling for a small test set would be necessary to establish the model's validity. 

The transfer learning approach provided better results with less training epochs than a non-transfer learning approach. 

\section{Conclusions}

The study has demonstrated the feasibility of creating training dataset for machine learning vision models employing openly available data as ground truth without the need for human manual labeling. The process is fast and easily adaptable to other regions, tasks or data. It is feasible to generate 
datasets for less common tasks, such as segmenting parking surfaces, 
given the availability of data. While OpenStreetMap (OSM) serves as a 
valuable source for ground truth in areas lacking comprehensive cartography, 
its incompleteness and errors can impede precise model outcomes in some 
instances. Despite inherent challenges related to the accuracy and completeness of ground truth data producing discrepancies between model predictions and ground truth, our models have shown sufficient accuracy and adaptability across diverse urban landscapes.

Validating models trained with incomplete data remains an open challenge. However, one potential solution is to test these models in a different city from where the training data was collected. This approach was applied in the case of parking models for Vienna and Madrid.

General purpose semantic segmentation models like SAM may lack 
precision for non-common tasks or images beyond the Western world. 
Therefore, the methodology developed in this study, using transfer learning with models like SAM, could contribute 
to the development of a larger amount of models suitable for a wider range of tasks and 
diverse environments. Transformer-based models, when employing pretrained 
encoders, outperform convolutional-based models with less training.

It was possible to detect temporal trends differentiating the tendencies in different areas in the city, contributing to the advancement of urban analytics. This research could help city administrations in assessing whether their policies to reduce car usage have effectively altered the city's land use and road and parking surface.

\subsection{Further research}

To enhance the dataset, the utilization of semi-supervised methods 
\cite{pelaez-vegas_survey_2023} could be employed, 
as these methods can help correct errors within the training dataset. In the training workflow examples where the models prediction differs vastly from the ground truth can be automatically excluded if the models prediction probability is high. 
Self and semi-supervised methods specifically developed for aerial 
image semantic segmentation can be explores to implement such workflows \cite{tang_semantic_2023}.
Accounting 
for class uncertainties during the labeling process can be factored 
into establishing weights in the loss function 
\cite{bressan_semantic_2022}. In order to give exact estimates for the models accuracy a small test set could be manually labelled if resources are available. \\

The models predictions could be used by official agencies to correct wrong data or manually supervise data where model and ground truth differ. Advancements in machine learning technology could motivate official agencies to establish more clear and accurate criteria when elaborating their cartography.

\section*{Acknowledgments}

Universidad Politécnica de Madrid (www.upm.es) provided  computing resources on the Magerit Supercomputer.

Miguel Ureña is supported by a contract funded by the Industrial Doctorates of the Community of Madrid (IND2020/TIC-17528 and IND2023/TIC-28743).

We want to thank Javier Sempere for proofreading this article.

\section*{Code}

The code developed for this paper is accessible on github: \\

Dataset creator and downloader: \url{https://github.com/GeomaticsCaminosUPM/GeoVisionDataset} \\

Machine learning model: \url{https://github.com/GeomaticsCaminosUPM/GeoVisionModels} \\

An example notebook to create a dataset: \url{https://github.com/GeomaticsCaminosUPM/GeoVisionDataset/blob/main/examples/vienna_transport_dataset.ipynb} \\

A notebook to try the SAM model: \url{https://github.com/GeomaticsCaminosUPM/GeoVisionModels/blob/main/examples/sam_text_based_segmentation_example.ipynb} \\

\newpage 

\bibliographystyle{IEEEtran}
\bibliography{references_bibtex.bib} 
\end{document}